\pdfoutput=1

\documentclass[11pt]{article}

\usepackage[]{acl}

\usepackage{comment}
\usepackage{rotating}
\usepackage[utf8]{inputenc}
\usepackage{times}
\usepackage{latexsym}
\usepackage{rotating,tabularx,siunitx}

\usepackage{float}
\usepackage{graphicx}
\usepackage{subfigure}
\usepackage{caption}
\usepackage{amsmath,amssymb}
\usepackage{mathtools}
\usepackage{booktabs}
\usepackage{multirow}
\usepackage{siunitx}
\usepackage{adjustbox}
\usepackage{pifont}

\usepackage{tabularray}

\usepackage[T1]{fontenc}

\usepackage[utf8]{inputenc}

\usepackage{microtype}
\usepackage{amsmath}
\usepackage{amsfonts}
\usepackage{multirow}

\usepackage{tabularx}
\usepackage{longtable}

\usepackage[ruled]{algorithm2e}
\usepackage{algpseudocode}
\usepackage{hyperref}
\usepackage[capitalise]{cleveref}
\usepackage{stfloats}
\usepackage{graphicx}
\usepackage{subfigure}
\usepackage{longtable}
\usepackage{makecell}
\usepackage{enumitem}
\usepackage{amssymb}
\usepackage{multirow}
\usepackage{color}
\usepackage{xcolor}
\usepackage{svg} 
\usepackage{diagbox}
\newtheorem{theorem}{Theorem}

\usepackage{bm}
\usepackage{defination}

\definecolor{spc}{RGB}{119, 107, 170}
\definecolor{pct}{rgb}{0.7, 0, 0.2}

\newcommand{\spc}{\textcolor{spc}{$\mathbf{\circ}$\,}}
\newcommand{\pct}{\textcolor{pct}{$\bullet$\,}}
\usepackage{multicol}

\title{\Large \bf \textsc{Centaur}: Bridging the Impossible Trinity of Privacy, Efficiency, and Performance in Privacy-Preserving Transformer Inference}

\author{{\bf Jinglong Luo}$^{1,2}$\quad{\bf Guanzhong Chen}$^{1}$\quad{\bf Yehong Zhang}$^{2}$\thanks{Corresponding author}\quad{\bf Shiyu Liu}$^{5}$ \\ {\bf Hui Wang}$^{2}$\quad{\bf Yue Yu}$^{2}$\quad{\bf Xun Zhou}$^{1,2}$\quad{\bf Yuan Qi}$^{3, 4}$\quad{\bf Zenglin Xu}$^{2,3,4*}$\\
$^{1}$Harbin Institute of Technology, Shenzhen, $^{2}$Pengcheng Laboratory\\
$^{3}$Fudan University, $^{4}$Shanghai Academy of AI for Science\\
$^{5}$Southwestern University of Finance and Economics\\
\texttt{\{luojl, zhangyh02\}@pcl.ac.cn}, \texttt{zenglin@gmail.com}
\\
}

\begin{document}
\maketitle
\begin{abstract}
With the growing deployment of pre-trained models like Transformers on cloud platforms, privacy concerns about model parameters and inference data are intensifying. Existing Privacy-Preserving Transformer Inference (PPTI) frameworks face the ``\textit{impossible trinity}'' of balancing privacy, efficiency, and performance: Secure Multi-Party Computation (SMPC)-based approaches ensure strong privacy but suffer from high computational overhead and performance losses; 
Conversely, permutation-based methods achieve near-plaintext efficiency and accuracy but compromise privacy by exposing sensitive model parameters and intermediate results.
Bridging this gap with a single approach presents substantial challenges, motivating the introduction of \textsc{Centaur}, a groundbreaking PPTI framework that seamlessly integrates random permutations and SMPC to address the ``impossible trinity''. By designing efficient PPTI algorithms tailored to the structural properties of Transformer models, \textsc{Centaur} achieves an unprecedented balance among privacy, efficiency, and performance. Our experiments demonstrate \textsc{Centaur}’s ability to resist diverse data reconstruction attacks, achieve plaintext-level inference accuracy, and boost inference speed by 5.0$\sim$30.4 times, unlocking new possibilities for secure and efficient AI deployment. 
\end{abstract}
\section{Introduction}
Transformer-based models~\cite{Vaswani-2017-Attention, bert, gpt2}, widely deployed in cloud services such as chatbots, virtual assistants, and code generators, have revolutionized many aspects of human activity. 
However, their cloud-based deployment introduces significant privacy risks. Companies deploying these models and users of the services must upload proprietary model parameters—critical to their competitive edge—along with potentially sensitive inference data, which could include personal information (e.g., identity, investment plans, or health records). These risks not only threaten the competitiveness of companies but also compromise individuals' privacy, raising concerns about whether cloud-based AI models can truly be trusted with sensitive information. Recently, Samsung banned its employees from using external large language model (LLM) services after an internal code leak\footnote{https://www.androidauthority.com/samsung-chatgpt-leak-3310307/}, further underscoring the growing privacy concerns.

\begin{figure}[t]
	\centering
\includegraphics[width= 0.45 \textwidth]{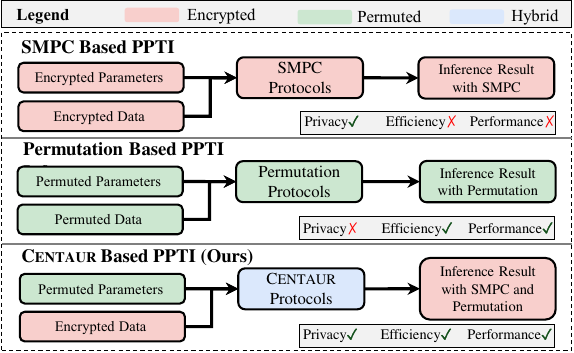} 
\caption{Overview of \textsc{Centaur} and Other PPTI Frameworks.}
\label{fig: Centaur_overview}
\end{figure}

Recent works \cite{hao2022iron,chen2022x,li2022mpcformer,luo2024secformer,yuan2023secure} have explored addressing the privacy concerns of model parameters and inference data in Transformer-based inference. However, these approaches often face the \textit{``impossible trinity'' of privacy, efficiency, and performance}. For example, SMPC-based privacy-preserving Transformer inference (PPTI) offers strong theoretical privacy guarantees but suffers from significant communication overhead. This inefficiency arises primarily from the \textit{numerous large-scale matrix multiplications and SMPC-unfriendly non-linear operations} inherent in Transformer models. To mitigate these issues, some studies \cite{li2022mpcformer,luo2024secformer} have replaced non-linear operations with linear ones, but this substitution results in further performance degradation (see \cref{sec: movation} for details).

In contrast, permutation-based PPTI \cite{yuan2023secure} achieves efficiency and performance comparable to plaintext inference by conducting plaintext computations on permuted model parameters and inference data. However, to ensure inference correctness, permutation-based PPTI must expose \textit{embedding layer parameters and some original intermediate results}, thereby introducing significant privacy leakage risks (see \cref{sec: movation} for details).

Existing PPTI frameworks struggle to balance privacy, efficiency, and performance, limiting their practical adoption in real-world applications. To bridge the “impossible trinity” and unlock new possibilities for secure and efficient AI deployment, we propose \textit{\textsc{Centaur}}, a practical PPTI framework that leverages the complementary strengths of multiple privacy-preserving strategies to protect the privacy of both model parameters and inference data (\cref{fig: Centaur_overview}). Specifically:

\begin{itemize}[leftmargin=* , itemsep=0pt]
\item \textbf{Privacy:} 
\textsc{Centaur} introduces a novel PPTI workflow, ensuring that model parameters, inference data, and intermediate results during inference remain either encrypted or in a randomly permuted state. The security analysis (\cref{sec: security}) and experimental results of data reconstruction attacks (\cref{sec: attack_exp}) demonstrate that \textsc{Centaur} effectively safeguards the privacy of both model parameters and inference data.

\item \textbf{Efficiency:} \textsc{Centaur} leverages random permutation to transform privacy-preserving multiplications between ciphertexts, which incur high communication overhead, into communication-free operations between plaintexts and ciphertexts, significantly improving the inference efficiency of \emph{linear layers} in PPTI. Additionally, it reduces the communication overhead of \emph{non-linear} operations in PPTI through the design of a series of privacy-preserving algorithms. Experimental results (\cref{sec: efficiency}) show that \textsc{Centaur} achieves inference speeds 5.0$\sim$30.4 times faster than existing SMPC-based PPTI frameworks.

\item \textbf{Performance:} 
\textsc{Centaur} preserves the original model structure and parameters by implementing precise computation of non-linear operations in Transformer models. Experimental results (\cref{sec: performance}) demonstrate that \textsc{Centaur} achieves performance identical to plaintext inference without the need for retraining or fine-tuning.
\end{itemize}

\section{Preliminaries}\label{sec: background}
\subsection{Transformer Models}\label{subsec: transformer models}
The Transformer model mainly consists of three components: the \textit{embedding layer}, the \textit{transformer layer}, and the \textit{adaptation layer}. In the embedding layer, the input features of the model are extracted as embeddings. At the transformer layer, the embedded information is processed through a multi-head attention mechanism and passed into the feed-forward neural network to produce a hidden state. In the adaptation layer, the hidden state is ultimately transformed into a vector representation that can be applied to various downstream tasks such as text classification and text prediction.

\subsection{Secure Multi-Party Computation} \label{sec: smpc}
Secure Multi-Party Computation (SMPC) enables a group of untrusted participants to jointly compute a function \(f\) without revealing private data. Among the various cryptographic primitives used to implement SMPC, secret sharing \cite{shamir1979share, GoldreichMW87} is widely employed in PPTI due to its efficiency. Specifically, 2-out-of-2 secret sharing divides a secret \(x\) in the integer ring \(\mathbb{Z}_L\) into two random shares \([\![x]\!] = ([x]_0, [x]_1)\), where neither share independently reveals any information about \(x\). The secret can be reconstructed by combining the shares as \(x = (([x]_0 + [x]_1) \mod L)\). In two-party SMPC protocols, these shares are distributed among \emph{two non-colluding parties}, who exchange masked intermediate results to perform privacy-preserving computations for various functions. At the end of the process, they each receive shares of the computed results.

\subsection{Permutation Matrix}\label{sec: premutation_matrix}
A permutation matrix \(\pi\) is a square matrix containing only \(0\)s and \(1\)s, with exactly one ``\(1\)'' in each row and column. In linear algebra, an \(n \times n\) permutation matrix represents a permutation of \(n\) elements and has the following key properties:  
\begin{itemize}[leftmargin=* , itemsep=0pt, labelsep=5pt]
    \item Multiplying a matrix by \(\pi\) permutes its rows (if \(\pi\) is on the left) or columns (if \(\pi\) is on the right).
    \item \(\pi\) is orthogonal, i.e., \(\pi \pi^\top = I\).
\end{itemize}
These properties make permutation matrices useful for privacy-preserving computations in Transformer models, enabling the following operations:  
\begin{itemize}[leftmargin=* , itemsep=0pt, labelsep=5pt]
    \item \textbf{Linear Layers:} For a linear layer with parameters \((W, B)\),
    \begin{equation}\label{eq: linear}
        Y = X\pi (W\pi)^\top + B = XW^\top + B.
    \end{equation}
    \item \textbf{Element-Wise Non-Linear Layers:} For an element-wise non-linear function \(f_e\),
   \begin{equation}\label{eq: non-linear}
    f_e(X \pi) = f_e(X) \pi.
    \end{equation}
\end{itemize}

The privacy offered by \(\pi\) increases with its size, making it ideal for large-scale Transformers. Specifically, an \(n \times n\) matrix has \(n!\) possible permutations. For example, when \(n = 1280\), the probability of brute-force recovery of the original matrix is approximately \(\frac{1}{1280!} \approx \frac{1}{2^{11372}}\).

\section{Impossible Trinity in PPTI}\label{sec: movation}

\paragraph{Observation 1: Efficiency and Performance Challenges of SMPC-Based PPTI.} SMPC-based PPTI can be formalized as a two-party SMPC protocol between the model developer and the client. In this setup, the shares of model parameters and inference data are used as inputs to the SMPC protocols, enabling privacy-preserving execution of Transformer operations.

This approach ensures privacy for model parameters and inference data but faces severe inefficiencies, primarily from the high communication overhead in large-scale matrix multiplications and non-linear operations within Transformers. For example, running BERT$_{\text{BASE}}$ inference with CrypTen \cite{knott2021crypten} in a WAN (200 Mbps bandwidth, 40 ms latency) takes 881 seconds, with 865 seconds spent on transmitting 66 GB of intermediate data.

Efforts to improve the efficiency of SMPC-based PPTI can be classified into two categories: 1) SMPC Protocol Design: Approaches such as \cite{hao2022iron, zheng2023primer, sigma, dong2023puma, chipherGPT, ding2023east, BLOT, lu2023bumblebee, luo2024secformer, li2024nimbus} focus on developing efficient privacy-preserving algorithms for non-linear operations in Transformers. While these methods preserve model performance, they still incur substantial computation and communication overhead.
2) Model Design: Techniques like \cite{li2022mpcformer, zeng2022mpcvit, SAL-VIT, liang2023merge} modify the model by replacing SMPC-unfriendly non-linear operations to reduce high computational overhead. Although these strategies improve efficiency, they often result in significant performance degradation (see \cref{tab: performance result} for details).

\begin{figure}[t]
	\centering
\includegraphics[width=0.48\textwidth]{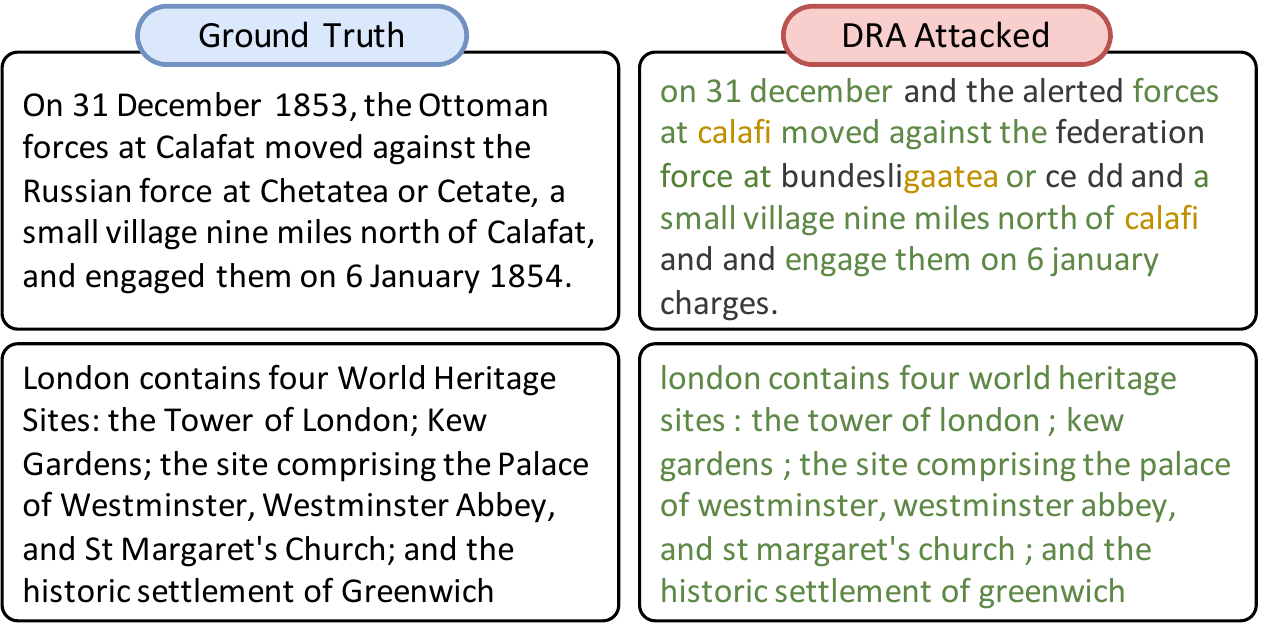} 
\caption{Two examples of recovering private inference input data through attacks on intermediate results. On the left are the real data, and on the right are the data reconstructed using data reconstruction attacks. Green indicates complete recovery, while orange signifies approximate recovery.}
\label{fig: atk_example_short}
 \vspace{-3mm}
\end{figure}

\paragraph{Observation 2: Privacy Leakage Risks in Permutation-Based PPTI.}\label{sec: random permutation leakage} 
Unlike SMPC-based PPTI, permutation-based PPTI uses permuted model parameters and inference data as input. By leveraging the properties of the permutation matrix, it correctly performs linear layers (matrix multiplication, \cref{eq: linear}) and nonlinear layers (element-wise operations, \cref{eq: non-linear}), producing permuted inference results. Since the computation is directly performed on the plaintext permuted data, permutation-based PPTI achieves efficiency and performance comparable to plaintext inference. However, it \emph{compromises the privacy of both model parameters and inference data}.

For model parameters, permutation-based PPTI faces the issue of sequence-level permutation vulnerability due to the relatively short length of the inference data sequence. \citet{yuan2023secure} suggest performing the permutation in the input feature space\footnote{The feature dimension \(d\) is typically large; for example, GPT-2$_{\text{LARGE}}$ has $d = 1280$.}. While this method enhances privacy, it requires the model developer to expose the \textit{embedding layer parameters} to the data owner.

Regarding inference data, the orthogonality of the permutation matrix (\cref{eq: linear}) leads to permutation-based PPTI \textit{revealing some original intermediate results}. We have demonstrated that existing data reconstruction methods can effectively recover the private inference data from these raw intermediate results. \cref{fig: atk_example_short} illustrates real examples of recovering the original data from the raw intermediate results, and more detailed attack results are provided in \cref{sec: attack_exp}.
\section{\textsc{Centaur}}\label{sec: centaur}
To bridge the ``\textit{impossible trinity}'' in PPTI, \textsc{Centaur} introduces a novel approach that seamlessly integrates random permutations and SMPC. This allows \textsc{Centaur} to overcome the limitations of existing methods, achieving a unique balance among privacy, efficiency, and performance.
The following sections delve into \textsc{Centaur}’s design and implementation. 

\subsection{Framework} \label{sec: overview}  
\begin{figure}[t]
	\centering
\includegraphics[width=0.48\textwidth]{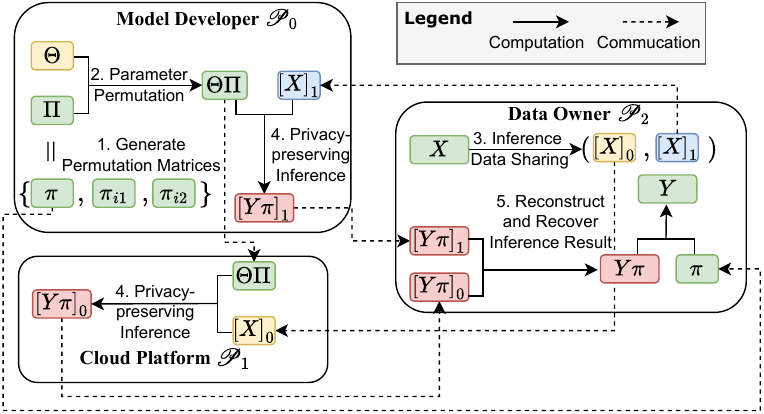} 
\caption{High-level Workflow of \textsc{Centaur}.}
\label{fig: centaur framework}
 \vspace{-3mm}
\end{figure}

\textsc{Centaur} focuses on the three-party scenario where the model developer and the cloud platform are separate entities, which is common in real-world model inference service providers \cite{yuan2023secure}. Specifically, as shown in \cref{fig: centaur framework}, \textsc{Centaur} involves three entities: model developer $\mathcal{P}_0$, cloud platform $\mathcal{P}_1$, and client $\mathcal{P}_2$. In this setup, $\mathcal{P}_0$ holds the private model parameters $\Theta$, while $\mathcal{P}_2$ holds the private inference data $X$.

\subsection{Threat Model} \label{subsec: threat model}
\textsc{Centaur} adopts the widely used three-party semi-honest model~\citep{Wagh2019SecureNN, ryffel2020ariann, li2022mpcformer, dong2023puma}. Specifically, it assumes that the model provider $\mathcal{P}_0$ does not collude with the cloud platform $\mathcal{P}_1$ to obtain the client $\mathcal{P}_2$’s private inference data, and likewise, the cloud platform $\mathcal{P}_1$ does not collude with the client $\mathcal{P}_2$ to access the model provider $\mathcal{P}_0$’s proprietary model parameters.

In contrast to two-party PPTI protocols~\citep{hao2022iron, BLOT, lu2023bumblebee}, which require the data owner to act as one of the computing parties and frequently communicate with the model provider during the entire inference process—often relying on homomorphic encryption or oblivious transfer to generate correlated randomness—\textsc{Centaur} takes a different approach. The data owner, who typically has limited computing and communication capabilities, only plays the role of a dealer, responsible for generating the correlated randomness required for PPTI execution, such as Beaver triples~\cite{beaver1992efficient} for multiplication and random permutation matrices to accelerate computation.

The computation and communication intensive tasks are delegated to the semi-honest cloud platform and the model developer, both of which are assumed to have ample resources. \textsc{Centaur} consists of two main phases: Initialization and Privacy-Preserving Inference, detailed as follows.

\vspace{2mm}
\noindent\textbf{Initialization.}\ \ The model developer $\mathcal{P}_0$ generates a set of random permutation matrices, $\Pi = \{\pi \in \mathbb{R}^{d \times d}, \pi_1 \in \mathbb{R}^{n \times n}, \pi_2 \in \mathbb{R}^{k \times k}\}$, where $n$ denotes the input length, $d$ represents the feature dimension, and $k$ corresponds to the intermediate dimension in the feed-forward neural network. These matrices are designed to permute the model parameters according to their respective dimensions. Among them, the permutation matrix $\pi$ is shared with $\mathcal{P}_2$. Subsequently, $\mathcal{P}_0$ applies the appropriate permutation matrix from $\Pi$ to permute the model parameters $\Theta$, resulting in the permuted parameters $\Theta^{\prime}$, which are then sent to $\mathcal{P}_1$.

\begin{figure}[ht]
	\centering
\includegraphics[width=0.48\textwidth]{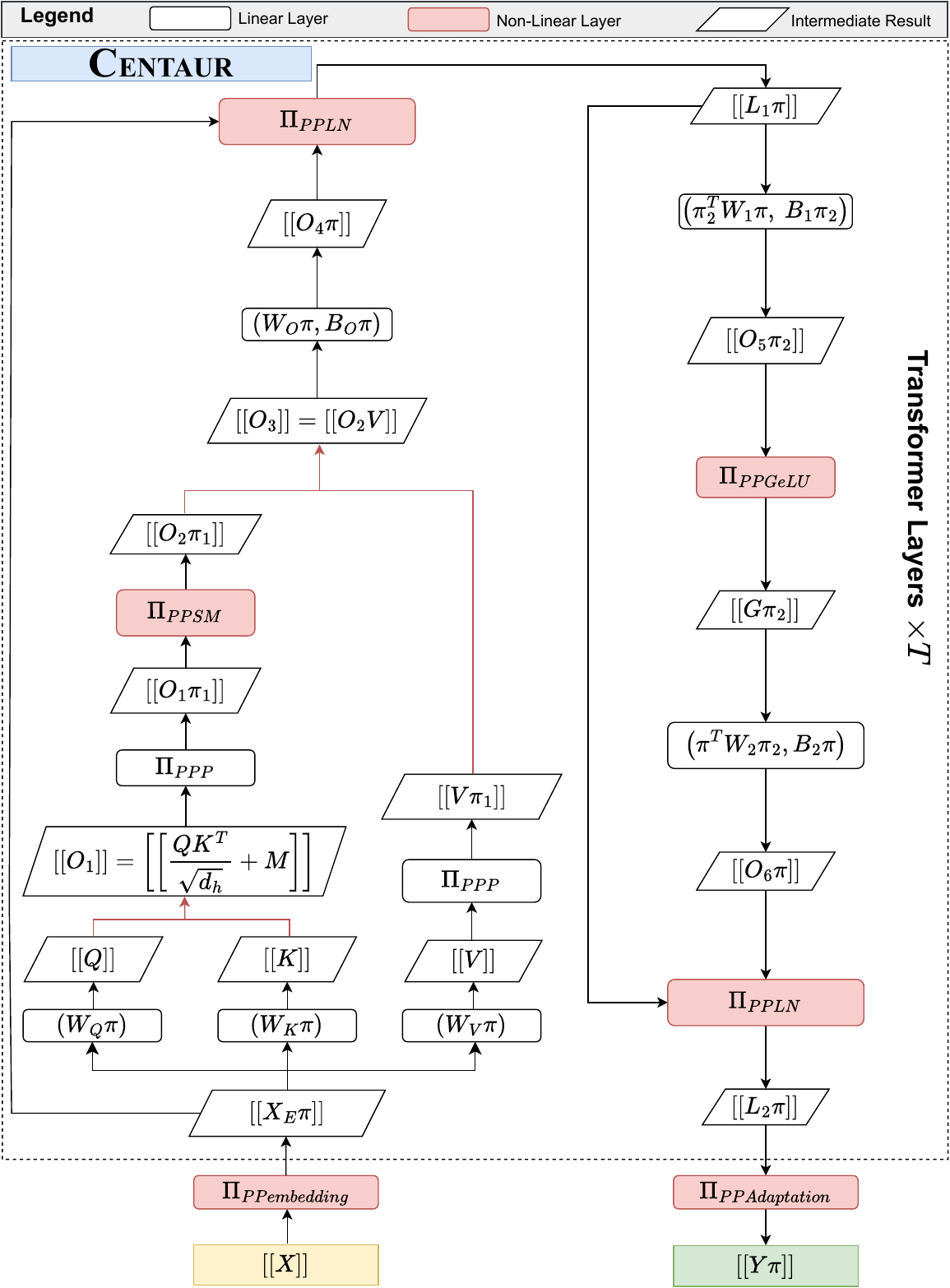} \caption{Implementation of \textsc{Centaur}-based PPTI. Red lines and boxes indicate that there is a communication overhead for the computation of this step. Black lines indicate completion of the calculation for that step without communication overhead.}
	\label{fig: centaur implementation}
    \vspace{-3mm}
\end{figure}

\vspace{2mm}
\noindent\textbf{Privacy-Preserving Inference.}\ \ The client $\mathcal{P}_2$ locally generates shares of the inference data $X \rightarrow ([X]_0, [X]_1)$ and sends $[X]_j$ to the respective parties $\mathcal{P}_j$ for $j \in \{0, 1\}$. Each $\mathcal{P}_j$ then takes $\Theta^{\prime}$ and $[X]_j$ as input, and jointly executes the privacy-preserving inference process according to the workflow shown in \cref{fig: centaur implementation}, resulting in the shares of the permuted inference result $[Y\pi]_j$. Subsequently, each $\mathcal{P}_j$ sends $[Y\pi]_j$ to the client $\mathcal{P}_2$. Upon receiving $[Y\pi]_j$, $\mathcal{P}_2$ reconstructs the permuted inference result $Y\pi = [Y\pi]_0 + [Y\pi]_1$, and restores the final inference result using $\pi$: $Y = Y\pi \pi^{\top}$.

\subsection{Implementation}\label{sec: centaur implementation}
As described in \cref{subsec: transformer models}, the Transformer model consists of the Transformer layers, the embedding layer, and the adaptation layer. We now outline how \textsc{Centaur} enhances privacy-preserving computation in each of these layers.

\subsubsection{Transformer Layers}\label{subsec: transformer_layer}
\vspace{2mm} \noindent\textbf{Linear Layer.}
\textsc{Centaur} optimizes the efficiency of linear layers by converting costly privacy-preserving matrix multiplications between random shares (denoted as $\Pi_{\text{MatMul}}$) into communication-free privacy-preserving multiplications between plaintexts and random shares (denoted as $\Pi_{\text{ScalMul}}$). This is achieved by separately using random permutation for model parameters and secret-sharing for inference data to ensure privacy.

As shown in \cref{fig: centaur implementation}, the linear layer parameters consist of ${W_Q, W_K, W_V, (W_O, B_O)}$ in the attention mechanism and ${(W_1, B_1), (W_2, B_2)}$ in the feed-forward neural network for a single Transformer block. During the initialization phase, these parameters are permuted by the model developer $\mathcal{P}_0$. When data, in the form of secret shares, passes through these linear layers, the computation is performed using the communication-free plaintext-shares privacy-preserving multiplication protocol $\Pi_{\text{ScalMul}}$. The shares of the computation results are then output as follows:
\begin{align}\label{eq: public mul}
[\![Q]\!] &= \Pi_{\text{ScalMul}}(W_Q \pi, [\![X_E \pi]\!]), \footnotemark \notag\\
[\![K]\!] &= \Pi_{\text{ScalMul}}(W_K \pi, [\![X_E \pi]\!]), \notag\\
[\![V]\!] &= \Pi_{\text{ScalMul}}(W_V \pi, [\![X_E \pi]\!]), \\
[\![O_4 \pi]\!] &= \Pi_{\text{ScalMul}}(W_O \pi, [\![O_3]\!]) + B_O \pi,\notag\\
[\![O_5 \pi_2]\!] &= \Pi_{\text{ScalMul}}(\pi_2^\top W_1 \pi, [\![L_1\pi]\!] ) + B_1 \pi_2,\notag\\
[\![O_6 \pi]\!] &= \Pi_{\text{ScalMul}}(\pi^\top W_2 \pi_2, [\![G\pi_2]\!] ) + B_2 \pi. \notag
\vspace{-2mm}
\end{align}
\footnotetext{We omit the bias here for concise presentation. For the case that there is additional bias parameters $B$ in producing $Q$, $K$, and $V$, the model developer can secretly share $B$ to cloud platform and add it to the output of $\Pi_{\text{ScalMul}}$ using $\Pi_{\text{Add}}$.}
To ensure the correctness and security of the inference results, \textsc{Centaur} requires a limited number of privacy-preserving matrix multiplications between shares in the attention mechanism. The detailed computation process is as follows:
\begin{equation}\label{eq: privacy mul}
\begin{split}
[\![O_1]\!] &= \Pi_{\text{MatMul}}([\![Q]\!], [\![K]\!])/\sqrt{d_h} + [\![M]\!], \\
[\![O_3]\!] &= \Pi_{\text{MatMul}}([\![O_2 \pi_1]\!], [\![V \pi_1]\!]).\\
\end{split}
\end{equation}
\vspace{-2mm}

\noindent\textbf{Non-linear Layers.} \label{subsec: element-wise}
\textsc{Centaur} enhances the efficiency of nonlinear layers by converting secret shares into a randomly permuted state, enabling plaintext computations for element-wise nonlinear operations on the permuted data.

For any nonlinear operation with permuted input \(X\pi\), which has been secret-shared between \(\mathcal{P}_0\) and \(\mathcal{P}_1\), the process proceeds as follows:
\begin{itemize}[leftmargin=* , itemsep=0pt]
\item The model developer \(\mathcal{P}_0\) sends the share \([X\pi]_0\) to the cloud platform \(\mathcal{P}_1\), enabling it to convert the input from the secret-sharing state \([\![X\pi]\!]\) to the permuted state \(X\pi\).
\item \(\mathcal{P}_1\) performs the nonlinear computation using \(X\pi\) and obtains the permuted output \(Y\pi\).
\item \(\mathcal{P}_1\) generates shares \([\![Y\pi]\!]\) of \(Y\pi\) and sends \([Y\pi]_0\) back to \(\mathcal{P}_0\).
\end{itemize}

This process requires two rounds of communication to transmit the shares of both the input and the output. Based on this, \textsc{Centaur} supports Privacy-Preserving Softmax (\(\Pi_{\text{PPSM}}\)), Privacy-Preserving GeLU (\(\Pi_{\text{PPGeLU}}\)), and Privacy-Preserving LayerNorm (\(\Pi_{\text{PPLN}}\)) for computing nonlinear layers in Transformers. Detailed construction algorithms are provided in \cref{sec: centaur algorithms}.

It is important to note that transitioning the input from the secret-sharing state \([\![X\pi]\!]\) to the permuted state \(X\pi\) requires the input shares to be in the permuted state. However, this condition is not always met in the PPTI process. For example, the shares of \(O_1\) are initially not in the permuted state because the permutation matrix \(\pi\) is canceled out during \(\Pi_{\text{MatMul}}\) (\cref{eq: privacy mul}). To address this, \textsc{Centaur} introduces a Privacy-Preserving Permutation (\(\Pi_{\text{PPP}}\)) protocol. By invoking privacy-preserving matrix multiplication, \(\Pi_{\text{PPP}}\) converts the shares of any input \([\![X]\!]\) into \([\![X \pi]\!]\). The detailed process is outlined in \cref{alg: ppp}.

\subsubsection{Embedding Layer \& Adaptation Layer.} 
The Embedding and Adaptation layers involve both linear and nonlinear operations, enabling dual acceleration of efficiency within \textsc{Centaur}. Specifically, the Embedding layer includes matrix multiplication and LayerNorm operations, allowing for Privacy-Preserving Embedding ($\Pi_{\text{PPEmbedding}}$) via the invocation of $\Pi_{\text{ScalMul}}$ and $\Pi_{\text{PPLN}}$. The construction of the Privacy-Preserving Adaptation ($\Pi_{\text{PPAdaptation}}$) layer, which adapts to different downstream tasks such as classification or prediction, varies across Transformer models. However, it can be uniformly implemented by using \textsc{Centaur}'s privacy-preserving algorithms. Detailed constructions for PPEmbedding and PPAdaptation are provided in \cref{alg: ppembedding} and \cref{alg: ppadaptation}.

\subsection{Theoretical Analysis}\label{sec: security}
\textsc{Centaur} can leverage the properties of permutation matrices to ensure the confidentiality of model parameters. By applying the widely used simulation-based paradigm \cite{Lindell2017simulate} from SMPC, we can demonstrate that intermediate results under secret-sharing can guarantee the confidentiality of user inference data. Additionally, using distance correlation theory \cite{discorr}, privacy protection of intermediate results under random permutation can be analyzed. We leave the detailed security analyses in \cref{sec: theoretical sec} since the theoretical framework of the SMPC and random permutation mechanism, which is usually the focus of the security community, is not over-explored by \textsc{Centaur}. We will focus on substantiating the \emph{empirical security} of \textsc{Centaur} through rich and complex attack experiments in \cref{sec: attack_exp}.

\begin{table*}[ht]
    \centering
    \resizebox{0.95 \linewidth}{!}{
    \begin{tabular}{cc|rrrrr||rrrrrr}
        \toprule
        &&  \multicolumn{5}{c||}{BERT$_{\text{LARGE}}$ on the QNLI dataset}
        &\multicolumn{4}{c}{GPT-2$_{\text{LARGE}}$ on the Wikitext-103 dataset}\\
        \cmidrule{3-12}
        Attacks & Methods     &\multicolumn{1}{c}{$O_1$}   &\multicolumn{1}{c}{$O_4$}      &\multicolumn{1}{c}{$O_5$}     
        &\multicolumn{1}{c}{$O_6$} &\multicolumn{1}{c||}{Avg}  
        &\multicolumn{1}{c}{$O_1$}   &\multicolumn{1}{c}{$O_4$}      &\multicolumn{1}{c}{$O_5$}      &\multicolumn{1}{c}{$O_6$}     &Avg
        \\
        \midrule
        \multirow{3}{*}{SIP}        &W/O        &$66.14\pm1.38$     &$78.64\pm0.28$ &$95.57\pm0.06$ &$96.00\pm0.05$   & $84.09$
        &$69.64\pm0.68$ &$92.91\pm0.17$ &$93.69\pm0.11$ &$94.31\pm0.21$ & $87.64$
        \\
        &W(Ours)          &$10.72\pm2.01$     &\underline{$2.03\pm0.89$}  &\underline{$0.00\pm0.00$} &\underline{$2.71\pm1.86$}  & \underline{$3.86$} &\underline{$6.10\pm4.67$}  &$12.90\pm0.64$ &$0.58\pm0.21$  &\underline{$2.00\pm1.29$} 
        & $5.40$\\
        &Rand       &\underline{$5.08\pm0.04$}      &$6.82\pm0.02$  &$0.17\pm0.06$  &$3.58\pm0.21$  & $3.91$ &$14.65\pm0.90$ &\underline{$2.69\pm0.05$}  &\underline{$0.00\pm0.00$} &$3.38\pm0.04$ 
        &\underline{$5.18$}\\
\midrule
        \multirow{3}{*}{EIA}        
        &W/O        &$100.00\pm0.00$     &$36.49\pm1.13$ &$80.97\pm0.71$  &$19.5\pm0.50$ & $ 59.24$ &$96.70\pm0.02$
        &$99.97\pm0.04$ &$100.00\pm0.00$ &$67.30\pm0.01$ & $ 90.99$
        \\
        &W(Ours)          &$1.37\pm0.12$     &\underline{$5.94\pm0.43$}  & $2.89\pm0.13$  & \underline{$0.12\pm0.07$} & $ 2.58$
        &$1.36\pm0.10$  &$11.90\pm0.37$  &$7.91\pm0.23$  &$4.40\pm0.33$ &
        $6.39 $\\
        &Rand       &\underline{$0.14\pm0.00$}      &$7.22\pm0.17$  &\underline{$0.34\pm0.11$}  &$0.85\pm0.03$   & \underline{$2.13 $}
        &\underline{$0.30\pm0.02$}  &\underline{$8.27\pm0.02$}  &\underline{$2.54\pm0.06$}  &\underline{$4.29\pm0.04$} &
       \underline{$3.85$} \\
        \midrule
        \multirow{3}{*}{BRE}        
        &W/O        &$56.64\pm1.06$     &$14.85\pm0.55$ &$74.50\pm0.75$  &$7.80\pm0.11$ & $38.45$ &$56.64\pm1.06$
        &$99.99\pm0.01$ &$99.99\pm0.00$ &$45.26\pm0.58$ & $75.47$
        \\
        &W(Ours)          &$0.21\pm0.02$     &$0.45\pm0.03$  &$0.52\pm0.39$  &\underline{$0.52\pm0.39$} & $0.43$
        &$0.21\pm0.02$  &$1.33\pm0.07$  &\underline{$0.03\pm0.01$}  &\underline{$0.07\pm0.02$} &
        $0.41$\\
        &Rand       &\underline{$0.07\pm0.02$}      &\underline{$0.25\pm0.20$}  &\underline{$0.09\pm0.01$}  &$0.58\pm0.02$   & \underline{$0.25$}
        &\underline{$0.07\pm0.02$}  &\underline{$0.20\pm0.00$}  &$0.08\pm0.00$  &$0.10\pm0.01$ &
       \underline{$0.11$} \\
        \bottomrule
    \end{tabular}
    }
 \caption{The degree of privacy leakage (ROUGE-L F1 Score (\%)) on the permuted intermediate results "$O_1, O_4, O_5, O_6$" using three data reconstruction attack methods. "W/O" represents the original data, "W" represents the permuted state, and "Rand" represents random input. The results denote the attack targets and are averaged over three different random seeds.}
    \label{tab: attack}
\end{table*}

\section{Experiments}\label{sec: experiments}
We conducted experiments to address three key questions regarding \textsc{Centaur}:
\textbf{Q1 (Privacy)}: Does the intermediate result in \textsc{Centaur}, stored in a randomly permuted state, withstand various rigorous adversarial attacks?
\textbf{Q2 (Efficiency)}: Can \textsc{Centaur} improve the inference speed of PPTI?
\textbf{Q3 (Performance)}: Does \textsc{Centaur} maintain the model's performance during PPTI execution?

\subsection{Experimental Setup}\label{sec: settings}
\noindent\textbf{Implementation.} We perform \textsc{Centaur} on CrypTen, a privacy-preserving machine learning framework based on SMPC. Our experiments were conducted on three servers, each equipped with an A100 GPU. To assess efficiency under varying conditions, we simulated different network settings using Linux Traffic Control. For the Local Area Network (LAN), the bandwidth was set to 3 Gbps with a round-trip delay of 0.8 ms, while for the Wide Area Network (WAN), the bandwidth was 100 Mbps with an 80 ms delay.

\noindent\textbf{Baselines.}  
\textsc{Centaur} is compared with several state-of-the-art PPTI frameworks: MPCFormer \cite{li2022mpcformer}, PUMA \cite{dong2023puma}, and SecFormer \cite{luo2024secformer}. MPCFormer improves PPTI efficiency by replacing Softmax and GeLU with SMPC-friendly quadratics. PUMA optimizes PPTI efficiency with enhanced SMPC protocols for nonlinear operations, while SecFormer also replaces Softmax with SMPC-friendly quadratics and refines protocols for nonlinear layers.

\begin{figure*}[ht]
	\centering
\includegraphics[width=0.99\textwidth]{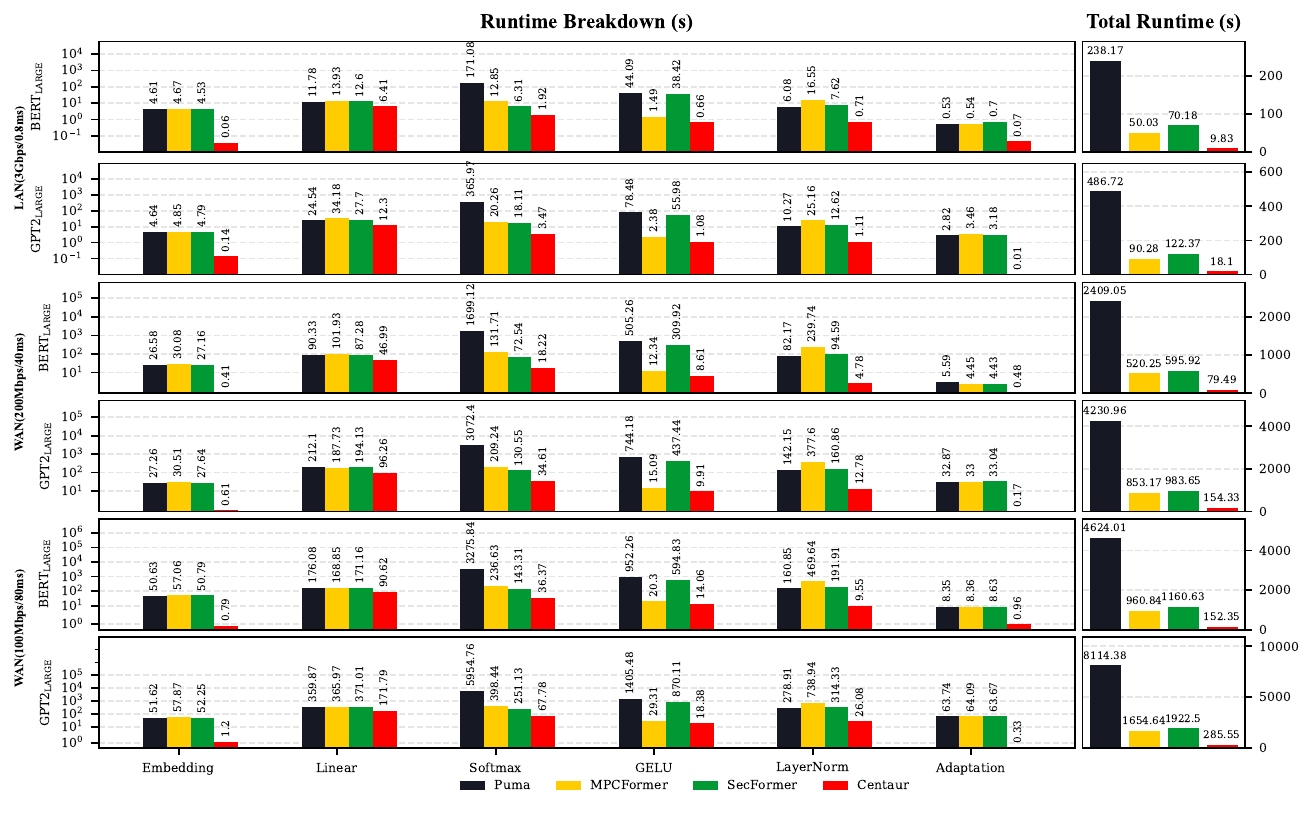} 
	\caption{Time breakdown for each operations (left) and the entire PPTI process (right) of the tested frameworks. The results are the average of ten runs.}
\label{fig: time_breakdown_large}
\end{figure*}

\noindent\textbf{Models and Datasets.}  
To ensure fairness, we selected the BERT and GPT-2 models, which are widely used in baseline evaluations, as benchmark models for our experimental assessment. For the BERT model, we selected five datasets from the GLUE benchmark \cite{GLUE} (RTE, CoLA, STS-B, MRPC, QNLI) for natural language understanding (NLU) tasks. For GPT-2, we employed two datasets from the Wikitext collection \cite{wikitext} (Wikitext-103 and Wikitext-2) for natural language generation (NLG) tasks. Furthermore, as \textsc{Centaur} performs the computation of nonlinear functions in Transformer models under permutation, it can theoretically be easily extended to other types of Transformer models, such as LLaMA, while maintaining a better balance between privacy, efficiency, and performance. These details will be further outlined in \cref{sec: generalizability}.

\subsection{Empirical Security}\label{sec: attack_exp}
To answer \textbf{Q1}, we conduct a series of rigorous adversarial experiments. Specifically, we first employ the three most advanced Data Reconstruction Attack (DRA) methods to attack the permuted intermediate results in an attempt to retrieve private inference data from users without recovering the permutation matrix. We also perform pattern-based and heuristic-based methods to recover the permutation matrix from the permuted intermediate results. The attack setup and more results of the attack experiments are presented in \cref{sec: empirical sec}.

\vspace{1mm}
\noindent\textbf{Attack Methods.} We evaluate three mainstream DRA methods targeting the intermediate outputs of Transformer models: (1) SIP \cite{bisr-ccs}, a learning-based approach that trains an inversion model on the auxiliary dataset to reconstruct the original sentence from any intermediate output derived from the private dataset; (2) Embedding Inversion Attack (EIA) \cite{eia-ccs}, an optimization-based approach that generates a dummy input and iteratively optimizes it (through relaxed optimization within the discrete vocabulary space) to match the observed intermediate outputs; and (3) BRE \cite{bisr-ccs}, an optimization-based approach that constructs dummy inputs but performs optimization within the continuous embedding space.

\vspace{1mm}
\noindent\textbf{Attack Targets.} 
In \textsc{Centaur}, as outlined in \cref{sec: centaur implementation}, intermediate results such as $O_1\pi_1$, $O_4\pi$, $O_5\pi_2$, and $O_6\pi$ are stored in permuted form on cloud platform $P_1$. To validate the privacy protection capabilities of \textsc{Centaur}, we conduct DRA experiments on these permuted results. For comparison, we also set up two control experiments: one with the original intermediate results ($O_1$, $O_4$, $O_5$, $O_6$), and another with random matrices of equivalent dimensions. The focus of our experiments is on the first Transformer block, where privacy leakage is most likely to occur.

\vspace{1mm}
\noindent\textbf{Evaluation Metrics.} We use ROUGE-L~\cite{rouge2004package}  F1 score as the evaluation metric for the attack experiments. ROUGE-L F1 assesses similarity based on the longest common subsequence, strictly following the order and tokens. By analyzing the ROUGE-L F1 values, we can understand the extent to which the original inference data can be reconstructed from the intermediate results. The ROUGE-L F1 score ranges from 0 to 1, with lower values indicating a lower recovery rate.

\vspace{1mm}
\noindent\textbf{Evaluation Results.} 
The experimental results in \cref{tab: attack} demonstrate that on both BERT$_{\text{LARGE}}$ and GPT-2$_{\text{LARGE}}$, the average ROUGE-L F1 values for data recovery by the three attack methods using \textsc{Centaur}’s permuted intermediate results are comparable to those obtained with random inputs. This further confirms that \textsc{Centaur} effectively preserves the privacy of inference data. Specifically, for BERT$_{\text{LARGE}}$, the three attack methods recover only $3.86\%$, $2.58\%$, and $0.43\%$ of the data's average ROUGE-L F1 values on the QNLI classification task dataset. In contrast, the recovery rate significantly increases when original intermediate results are used for the attacks. For instance, on GPT-2$_{\text{LARGE}}$, the average ROUGE-L F1 value for data recovery using EIA reaches as high as $90.99\%$, with $100\%$ data recovery achieved on $O_1=QK^{\top}$. This indicates that the privacy-preserving mechanism of PPTI \cite{yuan2023secure} based on random permutation completely fails once the original intermediate results are exposed.

\subsection{Efficiency Comparison} \label{sec: efficiency}
To address \textbf{Q2}, we analyze the inference time and communication overhead of \textsc{Centaur} performing PPTI and compare it with current state-of-the-art frameworks. The key results are presented in \cref{fig: time_breakdown_large}, with more details provided in \cref{sec: more efficiency results}. In two network settings—LAN (3Gbps, 0.8ms) and WAN (100Mbps, 80ms)—\textsc{Centaur} significantly outperforms other PPTI frameworks. For BERT$_{\text{LARGE}}$, \textsc{Centaur} is 5.1$\sim$24.2 times faster in a LAN environment and 6.3$\sim$30.4 times faster in WAN. For GPT-2$_{\text{LARGE}}$, \textsc{Centaur} is 5.0$\sim$26.9 times faster in LAN and 5.8$\sim$28.4 times faster in WAN. These efficiency improvements are attributed to \textsc{Centaur}'s dual optimization of both the linear and non-linear layers within PPTI.

\vspace{0.5mm}
\noindent\textbf{Linear Layers.} \textsc{Centaur} speeds up inference in linear layers by 1.8$\sim$2.2 times for BERT$_{\text{LARGE}}$ and 2.0$\sim$2.8 times for GPT-2$_{\text{LARGE}}$ compared to other PPTI frameworks. This is due to \textsc{Centaur}'s use of randomly permuted model parameters and secret-shared inference data, allowing most linear computations to be performed with the communication-free private matrix multiplication protocol $\Pi_{\text{ScalMul}}$.

\begin{table*}[ht]
\centering
\resizebox{0.99\linewidth}{!}{
    \begin{tabular}{lccccccc||cccccc}
    \toprule
    &QNLI (108k)    &CoLA (8.5k)    &STS-B (5.7k)   &MRPC (3.5k)    &RTE (2.5k)     &Avg.  
    &&Wikitext-2 (45k)  &Wikitext-103 (1800k)   &Avg.\\
    \hline
    &\multicolumn{6}{c}{BERT$_{\text{BASE}} ({\color{green}\uparrow})$ } &&\multicolumn{3}{c}{GPT-2$_{\text{BASE}} ({\color{red}\downarrow})$} \\
    \cmidrule{2-8}
    \cmidrule{9-11}
    Plain-text          
    &$91.7$ &$57.8$ &$89.1$ &$90.3$ &$69.7$ &\underline{$79.7$} &&$20.3$ &$24.3$ &\underline{$22.3$}\\
    PUMA                
    &$91.7$ &$57.8$ &$89.1$ &$90.3$ &$69.7$ &\underline{$79.7$} &&$20.3$ &$24.3$  &\underline{$22.3$}\\
    \pct MPCFormer$_{w/o}$    
    &$69.8$ &$0.0$  &$36.1$ &$81.2$ &$52.7$ &$48.0$ &&$420.9$   &$520.0$ &$470.5$\\
    \pct MPCFormer            
    &$90.6$ &$52.6$ &$80.3$ &$88.7$ &$64.9$ &$75.4$ &&$431.8$   &$522.3$ & $477.1$\\
    \spc SecFormer$_{w/o}$         
    &$89.3$ &$57.0$ &$86.2$ &$83.8$ &$63.2$ &$75.9$ &&$75.4$    &$131.0$ & $103.2$\\
    \spc SecFormer                 
    &$91.2$ &$57.1$ &$87.4$ &$89.2$ &$69.0$ &$78.8$ &&$75.3$    &$130.9$ &$103.1$\\
    \textbf{\textsc{Centaur}} (Ours)
    &$91.7$ &$57.8$ &$89.1$ &$90.3$ &$69.7$ &$\underline{79.7}$   &&$20.3$    &$24.3$ &$\underline{22.3}$\\

    \hline
    &\multicolumn{6}{c}{BERT$_{\text{LARGE}} ({\color{green}\uparrow})$} &&\multicolumn{3}{c}{GPT-2$_{\text{LARGE}} ({\color{red}\downarrow})$} \\
    \cmidrule{2-8}
    \cmidrule{9-11}
    Plain-text          
    &$92.4$ &$61.7$ &$90.2$ &$90.6$ &$75.5$ &\underline{$82.1$} &&$14.4$    &$16.0$ &\underline{$15.2$}\\
    PUMA                
    &$92.4$ &$61.7$ &$90.2$ &$90.6$ &$75.5$ &\underline{$82.1$} &&$14.4$    &$16.0$ &\underline{$15.2$}\\
    \pct MPCFormer$_{w/o}$    
    &$49.5$ &$0.0$  &$0.0$  &$81.2$ &$52.7$ &$36.7$ &&$94.4$    &$396.2$    &$245.3$\\
    \pct MPCFormer            
    &$87.8$ &$0.0$  &$52.1$ &$81.4$ &$59.2$ &$56.1$ &&$94.5$    &$402.5$    &$248.5$\\
    \spc SecFormer$_{w/o}$         
    &$90.8$ &$60.8$ &$89.0$ &$87.6$ &$69.7$ &$79.6$ && $91.8$   &$143.1$    &$117.5$\\
    \spc SecFormer                 
    &$92.0$ &$61.3$ &$89.2$ &$88.7$ &$72.6$ &$80.8$ && $91.5$   &$140.6$    &$119.1$\\
    \textbf{\textsc{Centaur}} (Ours)
    &$92.4$ &$61.7$ &$90.2$ &$90.6$ &$75.5$ &$\underline{82.1}$ &&$14.4$    &$16.0$     &$\underline{15.2}$\\
    \bottomrule
    \end{tabular}
}
    \caption{Performance comparison of BERT and GPT-2 models. Underlined numbers indicate the best results. Marker \spc refer to approximating GeLU with Quad. Marker \pct refer to approximating GeLU and Softmax with Quad and 2Quad, respectively. ``w/o'' indicates no re-training or knowledge distillation} 
    \label{tab: performance result}
\end{table*}

\vspace{1mm}
\noindent\textbf{Non-Linear Layers.} In the non-linear layers, \textsc{Centaur} achieves significant speed-ups. For Softmax and GeLU, \textsc{Centaur} outperforms the SMPC-based framework PUMA by two orders of magnitude. For BERT$_{\text{LARGE}}$, \textsc{Centaur} is 3.2$\sim$93.3 times faster in Softmax, 1.4$\sim$66.8 times faster in GeLU, and 8.6$\sim$50.1 times faster in LayerNorm. For GPT-2$_{\text{LARGE}}$, the speed-ups are 3.7$\sim$105.5, 1.5$\sim$76.5, and 9.3$\sim$29.5 times, respectively. These improvements are attributed to the privacy-preserving non-linear algorithms proposed in \textsc{Centaur}, which significantly reduce the communication overhead of non-linear computations in PPTI by converting the secret-share state to a random permutation state.

\vspace{1mm}
\noindent\textbf{Embedding \& Adaptation Layers.} The embedding and adaptation layers, which involve both linear and non-linear operations, benefit from \textsc{Centaur}’s dual optimization. For BERT$_{\text{LARGE}}$, \textsc{Centaur}’s inference speed in the embedding layer is 364.1$\sim$377.8 times faster, while for GPT-2$_{\text{LARGE}}$, the speedup ranges from 67.1$\sim$82.8 times. In the adaptation layer, \textsc{Centaur} accelerates BERT$_{\text{LARGE}}$ by 7.6$\sim$11.6 times and GPT-2$_{\text{LARGE}}$ by 193.7$\sim$290.9 times.

\subsection{Performance Comparison} \label{sec: performance}
To answer \textbf{Q3}, we validate the performance of \textsc{Centaur} and show the results in \cref{tab: performance result}.
As can be seen, both the BERT series models with an encoder structure and the GPT series models with a decoder structure achieve the same performance when using \textsc{Centaur} for PPTI as inference in plaintext. This indicates that \textsc{Centaur} does not compromise the performance of the plaintext models while protecting the model parameters and inference data. This is because \textsc{Centaur} does not make any adjustments to the structure of the plaintext Transformer models during the PPTI process. Consequently, \textsc{Centaur} can be combined with any existing Transformer architecture model to achieve PPTI with performance equivalent to plaintext inference.

\section{Discussion}
\textsc{Centaur} bridges the "impossible trinity" of privacy, efficiency, and performance in privacy-preserving transformer inference (PPTI) by leveraging the complementary strengths of SMPC and random permutation strategies. Comprehensive experiments demonstrate that \textsc{Centaur} significantly improves the efficiency of PPTI while providing a practical level of privacy, without sacrificing model performance. This enables \textsc{Centaur} to be readily integrated into existing Transformer-based model-as-a-service (MaaS) platforms to support privacy-preserving inference.

Moreover, \textsc{Centaur} does not yet incorporate other techniques designed to improve the computational and memory efficiency of Transformer inference, such as quantization and KV-cache, which could further enhance overall PPTI efficiency. These techniques are orthogonal to \textsc{Centaur}, and integrating them poses new challenges. For instance, KV-cache involves operations that are inherently incompatible with SMPC—such as similarity computation, top-k selection, and token aggregation—which would require additional considerations to enable privacy-preserving KV-cache. We leave the exploration of these directions for future work, aiming to further enhance the privacy and efficiency of PPTI by combining \textsc{Centaur} with such optimizations.

\section{Conclusion}\label{sec: conclusion}
This paper introduces \textsc{\textsc{Centaur}}, an efficient PPTI framework that employs tailored privacy-preserving mechanisms for both model parameters and inference data. By seamlessly integrating these techniques with customized algorithms, \textsc{\textsc{Centaur}} strikes an optimal balance in the privacy-efficiency-performance trade-off, often referred to as the ``\textit{impossibility triangle}'', unlocking new possibilities for the secure deployment of language models.

\section{Limitations}
\textsc{Centaur} adopts a privacy-preserving mechanism based on random permutation, which means that it cannot directly achieve theoretical security. \textsc{Centaur} does not overemphasize the theoretical security frameworks focused on the security domain but instead supports its claimed empirical security through extensive and complex attack experiments. In practical applications, privacy and usability are often incompatible. Particularly in the era of large models based on Transformer architectures, the rapid growth in model size has made traditional provable security techniques, such as SMPC and homomorphic encryption, impractical due to their high communication and computational costs. Therefore, we believe exploring practical privacy-preserving mechanisms for large models is of significant importance. Among various unverifiable security methods, the privacy-preserving capabilities of random permutation are positively correlated with the scale of the protected entity, making it especially suitable for large models with high-dimensional Transformer architectures. \textsc{Centaur} achieves a better balance between privacy and usability by combining random permutation with other provable security techniques. At the same time, we believe that the practical attack analyses on intermediate results in language models performed in \textsc{Centaur} hold equal importance to purely theoretical frameworks and require evaluation by the NLP community.

\section{Acknowledgements}\label{sec:Acknowledgements}
This work was partially supported by the National Natural Science Foundation of China (No. 62206139, No. 62472125, No. 72495122), the Natural Science Foundation of Guangdong Province, China (No. 2025A1515011258), and Shenzhen Sustained Support for Colleges \& Universities Program (No. GXWD20231128102922001), the
Major Key Project of PCL (PCL2023A09). 

\bibliography{ref}
\clearpage
\newpage
\clearpage
\appendix

\section*{\LARGE Appendices}
The appendices in this paper are organized as follows. 
\begin{itemize} [leftmargin=*]
    \item \cref{sec: centaur algorithms} presents the privacy-preserving algorithms designed in \textsc{Centaur}.
    \item \cref{sec: theoretical sec} offers a detailed theoretical security analysis of the \textsc{Centaur} framework.
    \item \cref{sec: empirical sec} offers a detailed empirical security analysis of the \textsc{Centaur} framework. 
    \item \cref{sec: more efficiency results} presents comparative analyses of \textsc{Centaur}'s communication volume and inference time for BERT$_{\text{BASE}}$ and GPT-2$_{\text{BASE}}$. 
    \item \cref{sec: generalizability} provides a comprehensive analysis of \textsc{Centaur} framework, further supported by experimental results on the LLaMA-7B model.
    \item Finally, \cref{sec: hyper-parameter} outlines the hyperparameters employed in the performance experiments.
\end{itemize}

\section{Privacy-preserving Algorithms in \textsc{Centaur}}\label{sec: centaur algorithms}
In this section, we present the construction of privacy-preserving algorithms within \textsc{Centaur}. Specifically, this includes Privacy-Preserving Softmax ($\Pi_{\text{PPSM}}$), Privacy-Preserving GeLU ($\Pi_{\text{PPGeLU}}$), Privacy-Preserving LayerNorm ($\Pi_{\text{PPLN}}$), Privacy-preserving permutation ($\Pi_{\text{PPP}}$), Privacy-Preserving Embedding ($\Pi_{\text{PPEmbedding}}$), and Privacy-Preserving Adaptation ($\Pi_{\text{PPAdaptation}}$). We illustrate the construction of $\Pi_{\text{PPAdaptation}}$ using the BERT series model as an example. In the BERT model, the Adaptation layer consists of a pooling layer composed of a linear layer $(W_P, B_P)$ and the activation function Tanh, followed by a linear layer with parameters $(W_C, B_C)$.

\begin{algorithm}
    \caption{Privacy-preserving Softmax ($\Pi_{\text{PPSM}}$)}
    \label{alg: ppsm}
    \LinesNumbered
    \SetNoFillComment
    \DontPrintSemicolon
    \KwIn{For $j \in \{0, 1\}$, $\mathcal{P}_j$ holds $[X \pi]_j$.}
    \KwOut{For $j \in \{0, 1\}$, $\mathcal{P}_j$ holds $[Y \pi]_j = [\text{Softmax}(X) \pi]_j$.}
The model developer $\mathcal{P}_0$ transmits $[X \pi]_0$ to $\mathcal{P}_1$\;
$\mathcal{P}_1$ reconstructs $X \pi$ and calculates $Y \pi = \text{Softmax}(X \pi) = \text{Softmax}(X) \pi$\;
$\mathcal{P}_1$ generates shares of $Y \pi$ and sends $[Y \pi]_0$ to $\mathcal{P}_0$
\end{algorithm}

\begin{algorithm}
    \caption{Privacy-preserving GeLU ($\Pi_{\text{PPGeLU}}$)}
    \label{alg:ppgelu}
    \LinesNumbered
    \SetNoFillComment
    \DontPrintSemicolon
    \KwIn{For $j \in \{0, 1\}$, $\mathcal{P}_j$ holds $[X \pi_2]_j$.}
    \KwOut{For $j \in \{0, 1\}$, $\mathcal{P}_j$ holds $[Y \pi_2]_j = [\text{GeLU}(X)\pi_2]_j$.}
The model developer $\mathcal{P}_0$ sends $[X \pi_2]_0$ to $\mathcal{P}_1$\;
$\mathcal{P}_1$ reconstructs $X \pi_2$ and calculates $Y \pi_2 = \text{GeLU}(X \pi_2)$\;
$\mathcal{P}_1$ generates shares of $Y \pi_2$ and sends $[Y \pi_2]_0$ to $\mathcal{P}_0$
\end{algorithm}

\begin{algorithm}
    \caption{Privacy-preserving LayerNorm ($\Pi_{\text{PPLN}}$)}
    \label{alg: ppln}
    \LinesNumbered
    \SetNoFillComment
    \DontPrintSemicolon
    \KwIn{For $j \in \{0, 1\}$, $\mathcal{P}_j$ holds $[X \pi]_j$.}
    \KwOut{For $j \in \{0, 1\}$, $\mathcal{P}_j$ holds $[Y\pi]_j = [\text{LayerNorm}(X) \pi]_j$.}
The model developer $\mathcal{P}_0$ transmits $[X \pi]_0$ to $\mathcal{P}_1$\;
$\mathcal{P}_1$ reconstructs $X \pi$ and calculates $Y \pi = \text{LayerNorm}(X \pi, \gamma \pi, \beta \pi)$\;
$\mathcal{P}_1$ generates shares of $Y \pi$ and sends $[Y \pi]_0$ to $\mathcal{P}_0$
\end{algorithm}

\begin{algorithm}
    \caption{Privacy-preserving Embedding ($\Pi_{\text{Embedding}}$)}
    \label{alg: ppembedding}
    \LinesNumbered
    \SetNoFillComment
    \DontPrintSemicolon
    \KwIn{For $j \in \{0, 1\}$, $\mathcal{P}_j$ holds $[X]_j$.}
    \KwOut{For $j \in \{0, 1\}$, $\mathcal{P}_j$ holds $[X_E \pi]_j$.}
$\mathcal{P}_0$ and $\mathcal{P}_1$ jointly calculate $[\![X_M \pi]\!] = \Pi_{\text{ScalMul}}([\![input]\!], W_E \pi)$\;
$\mathcal{P}_0$ and $\mathcal{P}_1$ jointly calculate $[\![X_E \pi]\!] = \Pi_{\text{PPLN}}([\![X_M \pi]\!])$
\end{algorithm}

\begin{algorithm}
    \caption{Privacy-preserving Adaptation ($\Pi_{\text{Adaptation}}$)}
    \label{alg: ppadaptation}
    \LinesNumbered
    \SetNoFillComment
    \DontPrintSemicolon
    \KwIn{For $j \in \{0, 1\}$, $\mathcal{P}_j$ holds $[X \pi]_j$.}
    \KwOut{For $j \in \{0, 1\}$, $\mathcal{P}_j$ holds $[Y \pi]_j$.}
$\mathcal{P}_0$ and $\mathcal{P}_1$ jointly calculate $[\![X_P \pi]\!] = \Pi_{\text{ScalMul}}([\![input]\!], W_P \pi)$\;
The model developer $\mathcal{P}_0$ sends $[X \pi]_0$ to $\mathcal{P}_1$\;
$\mathcal{P}_1$ reconstructs $X \pi$ and calculates $T \pi = \text{Tanh}(X \pi)= \text{Tanh}(X)\pi$\;
$\mathcal{P}_1$ generates shares of $T \pi$ and sends $[T \pi]_0$ to $\mathcal{P}_0$\;
$\mathcal{P}_0$ and $\mathcal{P}_1$ jointly calculate $[\![Y]\!] = \Pi_{\text{ScalMul}}([\![T \pi]\!], W_c)$
\end{algorithm}

\begin{algorithm}
    \caption{Privacy-preserving permutation ($\Pi_{\text{PPP}}$)}
    \label{alg: ppp}
    \LinesNumbered
    \SetNoFillComment
    \DontPrintSemicolon
    \KwIn{For $j \in \{0, 1\}$, $\mathcal{P}_j$ holds $[X]_j$.}
    \KwOut{For $j \in \{0, 1\}$, $\mathcal{P}_j$ holds $[X \pi]_j$.}
$\mathcal{P}_2$ generates a random permutation $\pi \in \mathbb{R}^{d\times d}$\;
$\mathcal{P}_2$ generates the shares $([\pi]_0, [\pi]_1)$ and sends $[\pi]_j$ to $\mathcal{P}_j$\;
$\mathcal{P}_0$ and $\mathcal{P}_1$ jointly calculate the permuated share $[\![X \pi]\!] = \Pi_{\text{MatMul}}([\![X]\!], [\![\pi]\!])$.
\end{algorithm}

\section{Theoretical Analysis}\label{sec: theoretical sec}
In this section, we theoretically demonstrate that \textsc{Centaur} can protect the confidentiality of both model parameters held by the model developer and user inference data. Specifically, we first leverage the properties of permutation matrices and the Transformer model structure to show how \textsc{Centaur} ensures the confidentiality of model parameters. Next, by applying the widely-used simulation-based paradigm from secure multi-party computation (SMPC), we illustrate how intermediate results in a secret-sharing state can safeguard the confidentiality of user inference data. Furthermore, we analyze the privacy-preserving capabilities of intermediate results under random permutation using distance correlation theory.

\subsection{Privacy of Model Parameters}\label{sec: parameter privacy}
In \textsc{Centaur}, the permutation matrices $\Pi = \{\pi, \pi_1, \pi_2\}$ are randomly generated locally by the model developer $\mathcal{P}_0$ during the initialization phase. Subsequently, $\mathcal{P}_0$ sends the permutation matrix $\pi$ to the client $\mathcal{P}_2$ and the permuted model parameters to the cloud platform $\mathcal{P}_1$. During the privacy-preserving inference phase, although $\mathcal{P}_1$ receives the permuted parameters in the linear layers and LayerNorm layers
$\{W_E\pi, W_Q\pi, W_K\pi, W_V\pi, (W_O\pi, B_O \pi), (\pi_2W_1\\\pi, B_1 \pi), (\pi W_2\pi_2,  B_2 \pi), (\gamma_1 \pi, \beta_1 \pi), (\gamma_2 \pi, \beta_2 \pi)\}$,
it lacks information about the permutation matrices $\{\pi \in \mathbb{R}^{d \times d}, \pi_2\in \mathbb{R}^{k \times k}\}$. This prevents $\mathcal{P}_1$ from directly obtaining the original parameters. Based on the properties of permutation matrices, the probability that $\mathcal{P}_1$ can derive the original parameters $\{W_E, W_Q, W_K, W_V, (W_O, B_O), (\gamma_1, \beta_1), (\gamma_2, \beta_2),\\ B_2\}$ from the permuted ones is $\frac{1}{d!}$. The probability of retrieving the parameters $\{W_1, W_2\}$ is $\frac{1}{d!k!}$ and $B_1$ is $\frac{1}{k!}$. 

Also, during both the initialization and privacy-preserving inference phases, the client $\mathcal{P}_2$ can only obtain the permutation matrix $\pi$ and the permuted inference results, thus preventing any access to information about the model parameters.

\subsection{Privacy of Inference Data}\label{sec: data privacy}
Unlike model parameters, inference data in \textsc{Centaur} is split into random shares. We prove that \textsc{Centaur} can ensure that during PPTI, neither the model developer \(\mathcal{P}_0\) nor the cloud platform \(\mathcal{P}_1\) can obtain any meaningful information about the inference data. Firstly, we prove through simulation that the intermediate results in the random shares state in \textsc{Centaur} do not leak the privacy of the inference data. Then, we demonstrate through distance correlation theory and various attack experiments to verify that the permuted intermediate results do not leak the privacy of the inference data.

\begin{table*}[ht]
    \centering
    \resizebox{0.95 \linewidth}{!}{
    \begin{tabular}{cc|rrrrr||rrrrrr}
        \toprule
        &&  \multicolumn{5}{c||}{BERT$_{\text{LARGE}}$ on the MRPC dataset}
        &\multicolumn{4}{c}{GPT-2$_{\text{LARGE}}$ on the Wikitext-2 dataset}\\
        \cmidrule{3-12}
        Attacks & Methods     &\multicolumn{1}{c}{$O_1$}   &\multicolumn{1}{c}{$O_4$}      &\multicolumn{1}{c}{$O_5$}     
        &\multicolumn{1}{c}{$O_6$} &\multicolumn{1}{c||}{Avg}  
        &\multicolumn{1}{c}{$O_1$}   &\multicolumn{1}{c}{$O_4$}      &\multicolumn{1}{c}{$O_5$}      &\multicolumn{1}{c}{$O_6$}     &Avg
        \\
        \midrule
        \multirow{3}{*}{SIP}   &W/O        &$70.94\pm0.17$     &$85.40\pm0.38$ &$97.61\pm0.08$  &$97.89\pm0.08$ & $87.96$ 
        &$65.38\pm0.14$ &$93.59\pm0.04$ &$93.07\pm0.13$ &$94.68\pm0.05$ & $86.68$
        \\
        &W(Ours)          &$10.96\pm1.24$     &\underline{$1.68\pm0.29$}  &\underline{$2.36\pm1.67$} &$4.96\pm0.67$ & $4.99$
        &$4.64\pm0.91$  &$11.58\pm0.47$  &$0.48\pm0.29$  &$2.68\pm2.71$ &
        $4.85$\\
        &Rand       &\underline{$5.72\pm0.05$}      &$6.09\pm0.04$  &$3.79\pm2.68$  & \underline{$4.14\pm0.19$}   & \underline{$4.94$}
        &\underline{$0.09\pm0.01$}  &\underline{$1.20\pm0.02$}  &\underline{$0.00\pm0.00$}  &\underline{$1.46\pm0.01$} &
       \underline{$0.69$} \\
        
\midrule
        \multirow{3}{*}{EIA}        
        &W/O        &$100.00\pm0.00$     &$34.25\pm0.62$ &$78.41\pm0.50$  &$19.31\pm0.78$ & $ 57.99$ &$96.17\pm0.05$
        &$100.00\pm0.00$ &$99.99\pm0.01$ &$65.04\pm 2.97$ & $ 90.30$
        \\
        &W(Ours)          &$1.60\pm0.40$     &\underline{$5.65\pm0.47$}  & $3.41\pm0.85$  & \underline{$0.25\pm0.21$} & $ 2.73$
        &$1.46\pm0.17$  &$12.49\pm0.25$  &$8.67\pm0.20$  &$4.89\pm0.77$ &
        $6.88 $\\
        &Rand       &\underline{$0.13\pm0.01$}      &$6.57\pm0.08$  &\underline{$0.28\pm0.01$}  &$0.77\pm0.03$   & \underline{$1.94 $}
        &\underline{$0.76\pm0.80$}  &\underline{$9.69\pm0.73$}  &\underline{$2.13\pm0.78$}  &\underline{$4.11\pm0.36$} &
       \underline{$4.17$} \\
        \midrule
        \multirow{3}{*}{BRE}   &W/O        &$51.89\pm1.26$     &$73.30\pm0.43$ &$70.86\pm0.37$ &$11.34\pm2.40$  & $51.85$
        &$100.00\pm0.00$ &$100.00\pm0.00$ &$100.00\pm0.00$ &$40.50\pm0.36$ & $85.13$
        \\
        &W(Ours)          &\underline{$0.07\pm0.01$}    &$2.77\pm0.11$  &$1.08\pm0.20$ &$0.91\pm0.36$  & $1.21$ &$0.26\pm0.14$  &$2.14\pm0.20$ &\underline{$0.04\pm0.01$}  &\underline{$0.07\pm0.01$} 
        & $0.63$\\
        &Rand       &$0.18\pm0.01$     &\underline{$1.94\pm0.08$}  &\underline{$0.68\pm0.03$}  &\underline{$0.54\pm0.05$}  & \underline{$0.84$} &\underline{$0.17\pm0.06$} &\underline{$0.28\pm0.07$}  &$0.06\pm0.02$ &$0.09\pm0.02$ 
        &\underline{$0.15$}\\     
        
        \bottomrule
    \end{tabular}
    }
 \caption{Attack performance (RougeL-F\%) on BERT$_{\text{LARGE}}$ and GPT-2$_{\text{LARGE}}$. The MRPC dataset is used for BERT and the Wikitext-2 dataset is used for GPT-2. ``W/O'' represents the original data without permutation; ``W'' represents the permuted state; ``Rand'' represents random input. Results are the average of three different random seeds.}
    \label{tab: more_att_result}
\end{table*}

\vspace{2mm}
\noindent\textbf{Intermediate Results in the Secret-Sharing State.} 
\textsc{Centaur} follows the semi-honest (also known as honest-but-curious) assumption, similar to \cite{li2022mpcformer, dong2023puma, luo2024secformer}. Under this assumption, the security of \textsc{Centaur} can be formally proven in the simulation paradigm, particularly against a static semi-honest adversary (denoted as \(\mathcal{A}\)). Specifically, the simulation paradigm divides the process into two distinct worlds: the real world and the ideal world. In the real world, the server executes the protocol in the presence of a semi-honest adversary \(\mathcal{A}\). In contrast, in the ideal world, the server transmits the input information to a trusted dealer who executes the protocol correctly. The security of the \textsc{Centaur} framework requires that the protocol executed with intermediate results in a randomly shared state produces distributions in the real world and the ideal world that are indistinguishable for any semi-honest adversary $\mathcal{A}$.

\begin{theorem}\label{th: th2}
 The protocols executed in \textsc{Centaur}, using intermediate results in a randomly shared state as input, satisfies the following criteria:
    \begin{itemize}[leftmargin=* , itemsep=0pt, labelsep=5pt]
        \item \textbf{Correctness:} For a model $F_{\Theta}$ with parameters $\Theta$ and inference data $X$, the output of the client at the end of the protocol is the correct inference result $F_{\Theta}(X)$.
        \item \textbf{Security:} For any corrupted computing server \(\mathcal{S}_j\) with \(j \in \{0,1\}\), there exists a probabilistic polynomial-time simulator \(Sim_{\mathcal{S}_j}\) such that the adversary \(\mathcal{A}\) cannot distinguish between \(View^{\Pi_{P}}_{\mathcal{S}_j}\) (i.e., the view of \(\mathcal{S}_j\) during the execution of \(\Pi_{P}\)) and \(Sim_{\mathcal{S}_j}\). 
    \end{itemize}   
\end{theorem} 

We provide the proof of \cref{th: th2} through the following analyses. 
According to \cref{fig: centaur implementation} and Eqs. \eqref{eq: public mul}-\eqref{eq: privacy mul}, the linear layers in a Transformer model only involve privacy-preserving operations $\Pi_{\text{PPP}}$ which is essentially a \(\Pi_{\text{MatMul}}\), $\Pi_{\text{ScalMul}}$, \(\Pi_{\text{MatMul}}\), and $\Pi_{\text{Add}}$. Since these basic operations $\Pi_{\text{ScalMul}}$, \(\Pi_{\text{MatMul}}\), and $\Pi_{\text{Add}}$ have been proven to satisfy \cref{th: th2}, we can directly prove that \textsc{Centaur} satisfies \cref{th: th2} for these linear layers using the universally composable security theorem established in \cite{canetti2001universally}.

\vspace{2mm}
\noindent\textbf{Intermediate Results in the Randomly Permuted State.} In \textsc{Centaur}, to perform non-linear operations such as $\Pi_{\text{PPSM}}$, $\Pi_{\text{PPGeLU}}$, and $\Pi_{\text{PPLN}}$, a conversion from a random sharing state to a random permutation state is required. During this process, the model developer $\mathcal{P}_0$ needs to send $[X \pi]_0$ to the cloud platform $\mathcal{P}_1$ for the reconstruction of $X\pi$, resulting in the intermediate results being in a random permutation state. 

We demonstrate both theoretically and experimentally that intermediate results in a random permutation state do not leak the privacy of inference data. Specifically, from a theoretical standpoint, we employ distance correlation theory \cite{discorr} to prove that the privacy leakage caused by intermediate results in a randomly permuted state is less than that of one-dimensional reduction, which has already been proven to possess privacy-preserving capabilities in practical applications \cite{wang2018theoretical,oliveira2004privacy}. According to \cite{zheng2022towards}, for any vector $o \in \mathbb{R}^{1 \times d}$, the following inequality holds:

\begin{equation}\label{eq: discorr}
\begin{aligned}
   \underset{\pi, W_A \in \mathbb{Z}^{d \times d}}{\mathbb{E}}& [\text{Discorr}(o, oW_A\pi)]\\
   \leq \underset{W_B \in \mathbb{Z}^{d \times 1}}{\mathbb{E}}& [\text{Discorr}(o, oW_B)], 
\end{aligned}
\end{equation}
where \text{Discorr} denotes a distance correlation function. This inequality implies that the distance correlation of the vector $o$ after passing through a linear layer with parameter $W_A$, followed by a permutation $\pi$, is less than or equal to the distance correlation after passing through a linear layer $W_B$ that compresses it to a 1-dimensional output. According to \cref{fig: centaur implementation}, all shares pass through at least one linear layer before being converted to a permuted state in \textsc{Centaur}. Therefore, it can be proven that the intermediate results in the permuted state in \textsc{Centaur} satisfy \cref{eq: discorr}.

\section{Empirical Security Analysis} \label{sec: empirical sec}

In this section, we demonstrate that the distributed secure inference based on the permutation of intermediate results provides \emph{empirical privacy security}. Specifically, in the scenario we consider, even for a reasonably strong attacker, the difficulty of successfully launching an attack is extremely high. To illustrate this, we assume an \emph{overly idealized} adversary, who has full white-box access to all parts of the model segment held by the model developer. This assumption is unrealistic in practical application scenarios. To comprehensively evaluate privacy, we further categorize the adversary into two types: those who launch attacks \emph{with and without cracking the permutation matrix}. It is important to note that, to date, no existing work has successfully compromised permuted Transformer intermediate results. We are the first to conduct a thorough analysis of the privacy and security of permutation-based Transformer inference. 

\vspace{0.2cm}
\noindent\textbf{Attack Setup.} We evaluate the privacy protection capabilities of \textsc{Centaur} by conducting a series of data reconstruction attack (DRA) experiments, with and without the adversary attempting to crack the permutation matrix (secret key). Consider an \emph{overly idealized attack scenario} where the adversary has unrestricted query access to key intermediate components of the model. An adversary can launch attacks at any nonlinear intermediate layer and recover the inference data's privacy using only the intermediate results from that layer.
Additionally, we assume this powerful adversary has access to an auxiliary dataset that may or may not resemble the target private dataset. We use a batch size of 4 and evaluate the average attack performance on 20 batches. To ensure the stability of the experimental results, each set of experiments was conducted with three different random seeds. The CNN-DailyMail News Text Summarization dataset~\cite{dataset-cnnnews}, which is entirely distinct from the target private datasets, was selected as the auxiliary dataset to simulate a realistic attack scenario.

\subsection{Attack without Cracking the Permutation Matrix}
For an attacker who does not attempt to crack the permutation matrix, the inability to determine whether the target intermediate results have been permuted leads them to employ traditional DRA strategies designed for the intermediate results of Transformers. In \cref{sec: attack_exp}, we have already provided experimental results for three state-of-the-art data reconstruction attack methods tailored for this scenario. Here, we present the attack setup and implementation details for the three adopted DRA methods, along with additional results and specific examples from the attack experiments discussed earlier.

\begin{figure*}[ht]
	\centering
\includegraphics[width=0.85\textwidth]{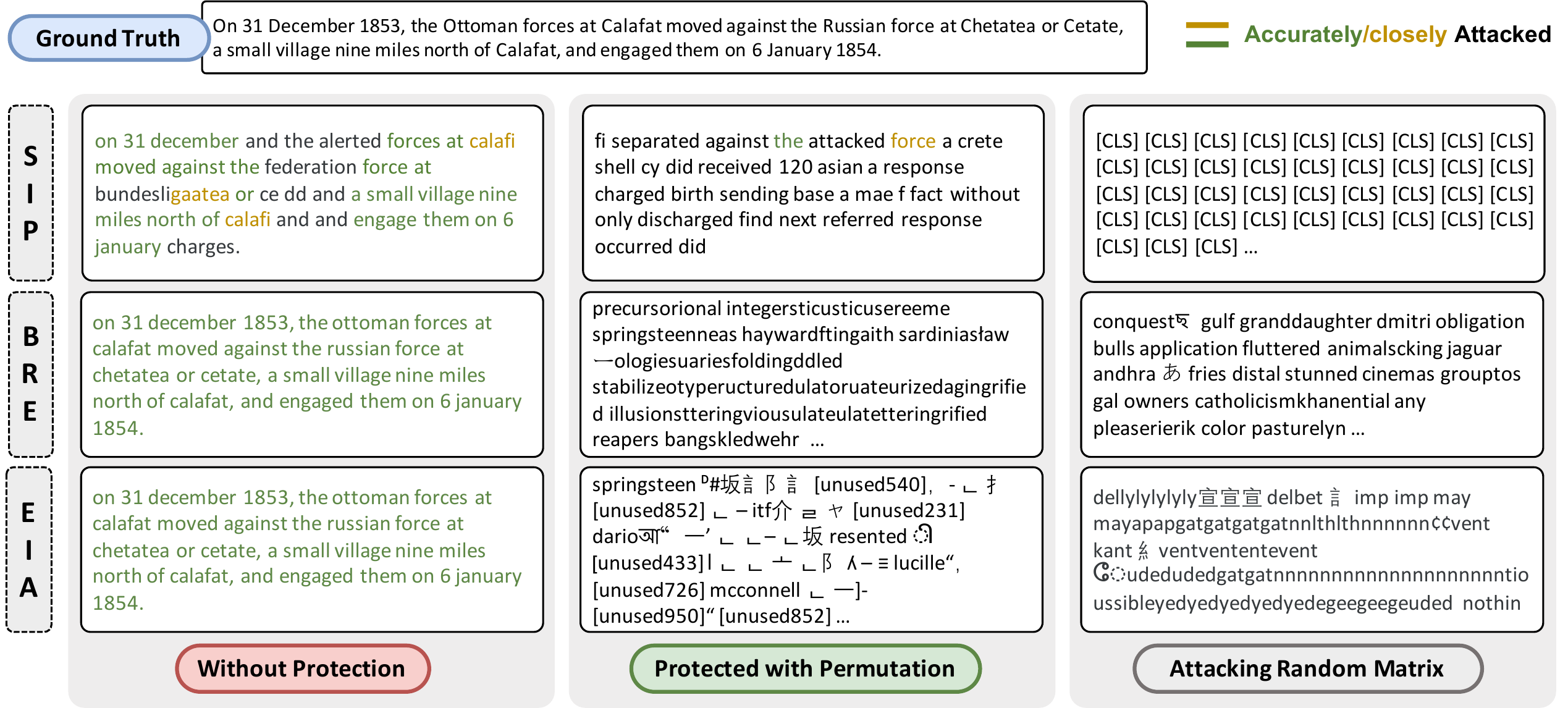} 
	\caption{An example of recovering private inference input data through $O_1$.}
\label{fig: atk_example_full}
\end{figure*}

\paragraph{Implementation Details. }For SIP, we employ a simple GRU model as the Inversion Model, with a hidden size of 256 and a dropout rate of 0.1, and train it for 20 epochs on the CNN Daily-Mail News dataset. Given that the last two dimensions of $O_1$ correspond to variable-length sequences, we truncate these sequences to a fixed length (512 in our experiments) before inputting them into the Inversion Model for training. For EIA, we use the Gumbel Softmax approximation to construct a distribution matrix over the vocabulary, which is then fed into the model. We optimize the intermediate outputs using Euclidean distance as the loss function. Since the attack focuses on intermediate results from the first layer, we do not need to apply the mapping strategy to shallow layers as described in ~\cite{eia-ccs}. For BRE, we directly construct an embedding, bypassing the embedding layer, and input it into the language model, optimizing based on cosine similarity. We conduct 6000 epochs of optimization for BRE and 2400 epochs for EIA, with both methods using AdamW with a learning rate of 0.1 as the optimizer.
\paragraph{More Attack Result. }We also report the outcomes of attacks on the MRPC dataset using the BERT$_{\text{LARGE}}$ model and on the Wikitext-2 dataset using the GPT-2$_{\text{LARGE}}$ model. Specifically, for the BERT$_{\text{LARGE}}$ model, the average ROUGE-L F1 scores for data recovery across three different attack methods on the MRPC classification task dataset are a mere $4.99\%$, $2.73\%$, and $1.21\%$, respectively. These results are comparable to the ROUGE-L F1 scores obtained when attacking random inputs. In contrast, attacks on plaintext intermediate results yield significantly higher recovery rates. Notably, the average ROUGE-L F1 score for data recovered using SIP from plaintext intermediate results reaches as high as $87.96\%$. A similar pattern is observed with the GPT-2$_{\text{LARGE}}$ model during prediction tasks. On the Wikitext-2 dataset, the average ROUGE-L F1 scores for data recovery from randomly permuted intermediate results are $4.58\%$, $6.88\%$, and $0.63\%$, which are again comparable to the recovery rates from random inputs. However, when targeting plaintext intermediate results, the average ROUGE-L F1 scores for data recovery using the three attack methods are significantly higher, with the EIA method recovering over $90.3\%$ of the private data.

\paragraph{Attack Examples. } We provide additional practical attack examples targeting $O_1 = QK^{T}$. These examples clearly demonstrate that directly attacking the plaintext $O_1$ can effectively recover private inference data, indicating that permutation-based PPTI presents a significant privacy leakage risk. In contrast, attacking obfuscated intermediate results or random inputs only produces meaningless garbled output. This demonstrates that the privacy protection provided by \textsc{Centaur} can effectively resist current DRA attacks.

\begin{figure}
    \centering
\includegraphics[width=\linewidth]{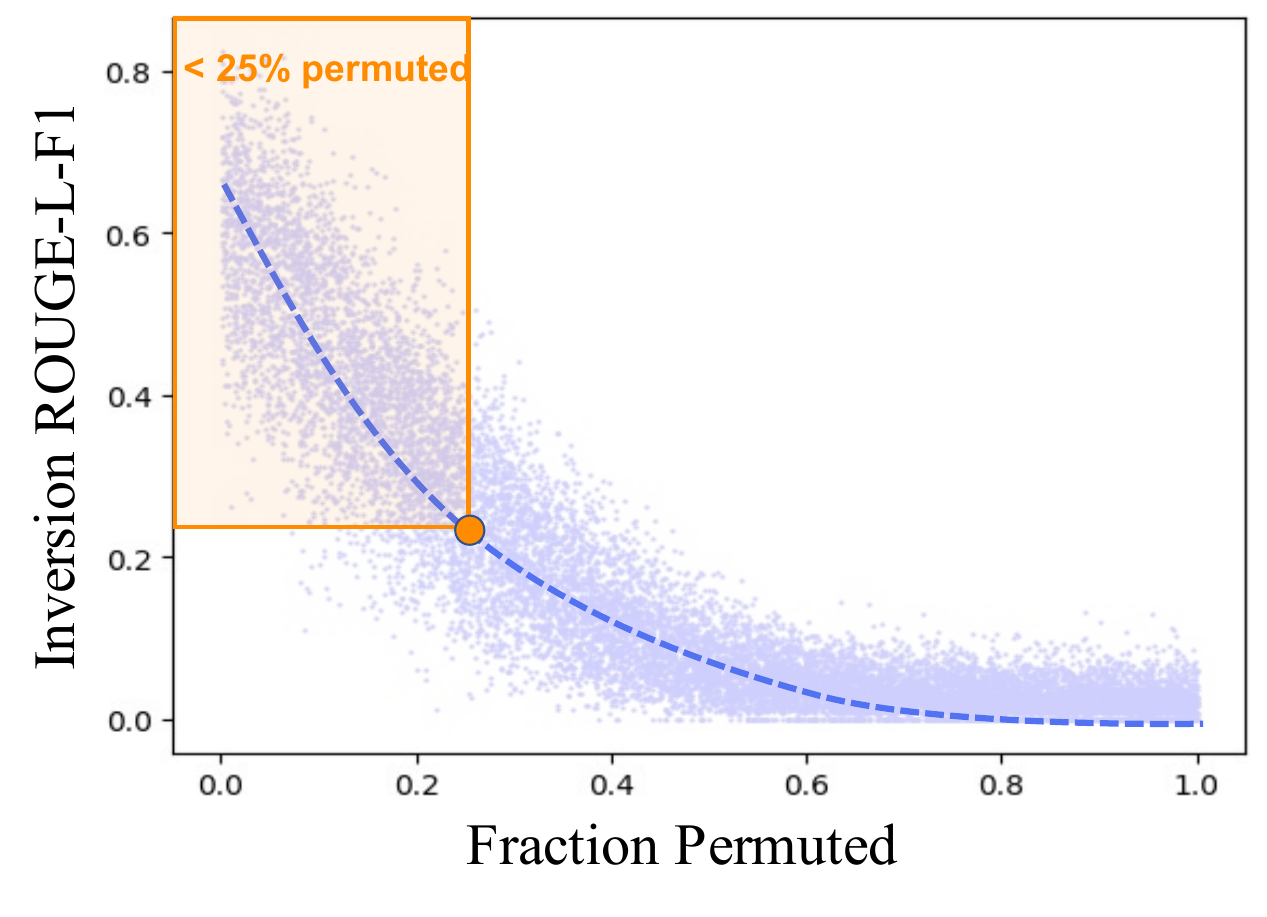}
    \caption{The correlation between the proportion of shuffled features and the effectiveness (measured by the ROUGE-L F1 score) of the inversion attack \( f^{-1} \), showing that reconstructing the raw text requires 75\%+ of the features to remain in place.}
    \label{fig:inv_frac}
\end{figure}

\paragraph{Analysis.} For the considered data reconstruction attacks, to launch an attack based on observations in the intermediate space \( N \), the attacker must obtain an inverse mapping \( f^{-1} \) to map the results back to the vocabulary space \( V \). In this context, we investigate the correlation between the proportion of shuffled features in the intermediate results and the effectiveness of \( f^{-1} \). The results, after fitting and smoothing, are presented in \cref{fig:inv_frac}. It is evident that a small amount of feature displacement (20\%) can significantly reduce the effectiveness of \( f^{-1} \). In practice, for the permutation matrix generated by np.permutation, when the hidden size of the large language model (LLM) exceeds 768, the proportion of non-shuffled elements is less than 0.13\%, effectively achieving near-complete feature reordering. This leads to the complete disruption of \( f^{-1} \). Thus, although the intermediate representations of the Transformer are sparse, in a scenario where almost all features are randomly reordered, \emph{any direct attack method that does not consider cracking the permutation matrix is impractical}.

\subsection{Attack by Cracking the Permutation Matrix}
Furthermore, we consider a more advanced adversary, who is aware that the intermediate result being attacked has been permuted and attempts to launch an attack by cracking the permutation matrix. We emphasize that permuting the intermediate results of a Transformer is \emph{difficult to crack in practice}, which stems from: 

\begin{itemize}
    \item\textbf{Huge secret key space}: A typical Transformer model has a large dimensionality for its intermediate representations. For instance, the dimensionality is often 768 (and it is even over a thousand for GPT-2 and Llama). The key space reaches $768!$. Even for a computer with a computing power of $10^{18}$ FLOPS, it is impossible to solve the problem within a reasonable time frame. 

    \item\textbf{Noisiness of intermediate representations}: Usually, the cracking of substitution ciphers is carried out directly in the vocabulary space $V$. However, in the context of Centaur, the target model $f:V\rightarrow N$ maps the original sentences to the intermediate space $N$. For the attacker under consideration, the cracking process occurs in the space $N$. For an attacker aiming to reconstruct the original sentence data from a target, \emph{the function $f$ is noisy}. Due to the stacking of attention mechanisms, the intermediate activations of the same token vary across different contexts and also differ from the initial embedding of that token. That is to say, the randomness here comes from the context during inference. 
\end{itemize}

Due to the adversary's limited attack view caused by the perturbation, the large key space, and the challenging inversion curve shown in \cref{fig:inv_frac}, cracking the permutation matrix proves to be difficult in practice. In the following, we attempt two cracking methods, namely pattern-based and searching-based approaches, both of which fail to successfully break the permutation matrix.

\subsubsection{Crack by Pattern Identification: Difficult}

For permutation-based encryption in the feature space, a key issue is whether there are identifiable patterns across the feature dimensions that could be exploited by the attacker to launch a cracking attack. We note that, due to operations such as LayerNorm performed by the Transformer on the feature dimensions, it is difficult to attempt cracking by simply identifying patterns in the different features, as \emph{their distributions are too similar to be distinguished}.

\begin{table}[h]
\caption{The average global Jensen-Shannon (JS) divergence across the feature dimensions of the intermediate results generated during inference on different models and datasets, with all values being less than 0.1, indicates that the distribution differences across the feature dimensions are minimal.}
\label{tab:js}
\resizebox{\columnwidth}{!}{%
\begin{tabular}{l|llll}
\hline
Model & PIQA & WikiText & MRPC & QNLI \\ \hline
BERT-large & $0.0748$ & $0.0618$ & $0.0605$ & $0.0612$ \\
GPT2-large & $0.0555$ & $0.0441$ & $0.0462$ & $0.0470$ \\ \hline
\end{tabular}%
}
\end{table}

\paragraph{Distribution Similarity Test} We calculated the global Jensen-Shannon (JS) divergence (ranging from 0 to 1, where 1 indicates a clear distinction between distributions) among all feature dimensions of the intermediate activations on BERT, GPT-2, and three datasets. It can be observed from \cref{tab:js} that all the global JS divergences are less than 0.1. The differences in the distributions of different features are extremely small. Moreover, considering that the intermediate dimension is quite large (>= 768), it is very difficult in practice to recover the permutation matrix by observing the distributions of these features.

\paragraph{Classifier-based Test}

We also attempted to use RNN and Linear as classifiers to model the distribution characteristics of different feature dimensions. However, even after careful tuning, such classifiers failed to fit successfully during the training process.

\subsubsection{Crack by Heuristic Searching: Difficult}
Cracking strategies that use heuristic signals such as frequency as search guides are indeed efficient in traditional substitution cipher scenarios. However, in the scenario considered by Centaur, \emph{the presence of noise makes it difficult for attackers to find effective and accurate heuristic signals}.

\begin{table*}[]
\caption{The attack performance (measured by ROUGE-L F1 score \%) of heuristic-searching-based cracking after the search curve reaches saturation, using different heuristic signals.}
\resizebox{\textwidth}{!}{%
\begin{tabular}{l|lllll}
\hline
{ \textbf{Heuristic Signals}} & {\textbf{\begin{tabular}[c]{@{}l@{}}Scoring-\\ model\end{tabular}}} & {\textbf{\begin{tabular}[c]{@{}l@{}}Frequency\\ (top1-token)\end{tabular}}} & {\textbf{\begin{tabular}[c]{@{}l@{}}Frequency\\ (top5-token)\end{tabular}}} & {\textbf{\begin{tabular}[c]{@{}l@{}}(GT)\\ Edit Distance\end{tabular}}} & {\textbf{\begin{tabular}[c]{@{}l@{}}(GT)\\ Invertion Rouge\end{tabular}}} \\ \hline
{Genetic Algorithm} & {$1.02\pm 0.98$} & { $8.23\pm 2.44$} & {$14.54\pm 3.84$} & {$11.45\pm 1.01$} & {$28.41\pm 1.98$} \\
{Gradient-based} & { $6.87\pm3.21$} & { $7.85\pm 2.90$} & { $9.87\pm 1.92$} & { -} & { $18.23\pm 2.09$} \\ \hline
\end{tabular}%
}
\label{tab:hs}
\end{table*}

Take the frequency-based attack as an example. Different from the monoalphabetic substitution cipher, the substitution space (768!) and the vocabulary space (>10000) are much larger than the alphabet. Moreover, the "substitution" occurs in the intermediate results rather than the original vocabulary, even if an attacker might obtain the intermediate representation of a known token, they still cannot directly solve for the permutation matrix as in a known-plaintext attack (KPA), because there is a random perturbation between the intermediate representation they possess and the one they observe. 

We conducted experiments to further verify the difficulty of heuristic search attacks. We consider both genetic algorithm~\citep{DBLP:conf/gecco/BassinB20} and  gradient-based continuous approximation approaches for searching the permutation matrix. We tried various heuristic schemes to guide the cracking process, including:
\begin{itemize}
    \item{Frequency-based.} The attacker can use the clustering of intermediate results (since permutation does not affect clustering based on metrics such as cosine-similarity) to count token frequencies. After identifying high-frequency tokens, the attacker can crack the permutation matrix by comparing the intermediate representations of high-frequency tokens before and after permutation. In the experiment, we assumed that the attacker had completely determined the identities of the top-5 and top-1 high-frequency permuted intermediate results. We attempted to use the cosine similarity between the original intermediate results of these five tokens (sampled by the attacker from the auxiliary dataset) and the observed values as a heuristic signal.
    \item{Scoring-model-based.} An ML model can be used to model the relationship between the "degree of disorder" and the degree of restoration mentioned in \cref{fig:inv_frac}. This model takes the permuted intermediate results as input and can score the degree of disorder of the permutation. In practice, a scoring model with the architecture of the bert - base model can fit this relationship. Therefore, an attempt is made to use the output score of this model as a heuristic signal.
    
    \item{Control.} As a control, we used two \emph{ground truth} metrics: (a) the mean edit distance (the average edit distance between the cracked permutation matrix and the real permutation matrix), and (b) ROUGE-L of the reconstructed sentence after cracking, compared to the real sentence, as ground truth heuristic signals.
    
\end{itemize}

We conducted permutation cracking experiments on an experimental machine equipped with 2 x Intel Xeon Gold 2.60GHz CPUs and 4 x NVIDIA A100(40GB) GPUs. We also recorded the ROUGE-L of the decrypted results of the cracked permutation after the search curve reached saturation (i.e., after the heuristic indicators stopped increasing for a certain period of time). It can be seen from \cref{tab:hs} that even when using the ground truth (GT) as the heuristic signal for the search, this search task remains difficult (it's hard to break through the 30\% performance bottleneck). As mentioned in 2.1, an attacker needs to recover more than 80\% of the permutation matrix to achieve an attack performance of over 30\%, which is already extremely challenging.  Moreover, the heuristic signals adopted by the attacker are perturbed. This perturbation will further confound the features that are already difficult to distinguish as mentioned in \cref{fig:inv_frac}, creating an inevitable gap between them and the real signals. This further prevents the recovery performance from surpassing the 30\% bottleneck.

\section{More Efficiency Results}\label{sec: more efficiency results}
\begin{figure*}[!ht]
	\centering
\includegraphics[width=0.99\textwidth]{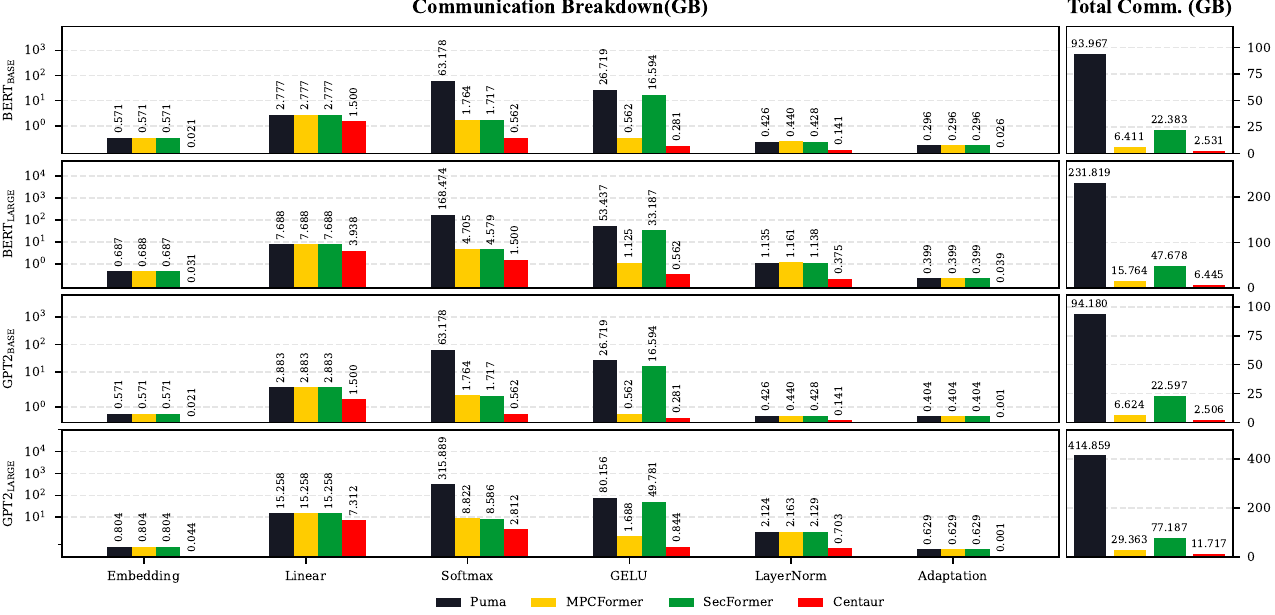} 
	\caption{Communication volume for each operations (left) and the entire PPTI process (right) of the tested frameworks.}
\label{fig: comm_breakdown}
\end{figure*}
\subsection{Communication Overhead Analyses} \label{sec: comm_overhead}
We analyze the communication overhead of \textsc{Centaur}-based PPTI and compare it with the current leading privacy-preserving inference frameworks. For BERT$_\text{BASE}$ and BERT$_\text{LARGE}$, using \textsc{Centaur} for PPTI reduces the communication overhead, respectively, by 2.5 $\sim$ 37.1 and 2.4$\sim$36.0 times compared to existing methods. For the GPT-2$_\text{BASE}$ and GPT-2$_\text{LARGE}$, this reduction is 2.6$\sim$37.6 and 2.51$\sim$35.4 times, respectively. This significant reduction is attributed to the hybrid computation mechanism employed by \textsc{Centaur}, which drastically reduces the communication overhead in both the linear and non-linear layers during PPTI.

\vspace{1mm}
\noindent\textbf{Linear Layers.} In the linear layers, the communication overhead required for performing PPTI using \textsc{Centaur} is half of existing PPTI frameworks. This is because in the baseline PPTI frameworks, both the model parameters and inference data are in secret-sharing states, requiring the use of the private matrix multiplication protocol $\Pi_{\text{MatMul}}$ between secret shares during linear layer operations. In contrast, \textsc{Centaur} places only the inference data in a secret-sharing state while keeping the model parameters in a randomly permuted state. This allows \textsc{Centaur} to perform most of the linear layer computations using the communication-free private matrix multiplication protocol $\Pi_{\text{ScalMul}}$ between plaintext and secret shares.

\vspace{1mm}
\noindent\textbf{Non-Linear Layers.} In the non-linear layers, \textsc{Centaur} significantly reduces the communication overhead of privacy-preserving computations by converting between secret-sharing and random permutation states. Specifically, for the privacy-preserving computation of Softmax, \textsc{Centaur} reduces the communication overhead by 3.1$\sim$112.3 times compared to the current state-of-the-art PPTI frameworks. For the privacy-preserving computation of GeLU, \textsc{Centaur} reduces the communication overhead by 2.0$\sim$95.0 times, and for LayerNorm, \textsc{Centaur} reduces the communication overhead by 3.0 $\sim$ 3.1 times.

\begin{figure*}[!ht]
	\centering
\includegraphics[width=\textwidth]{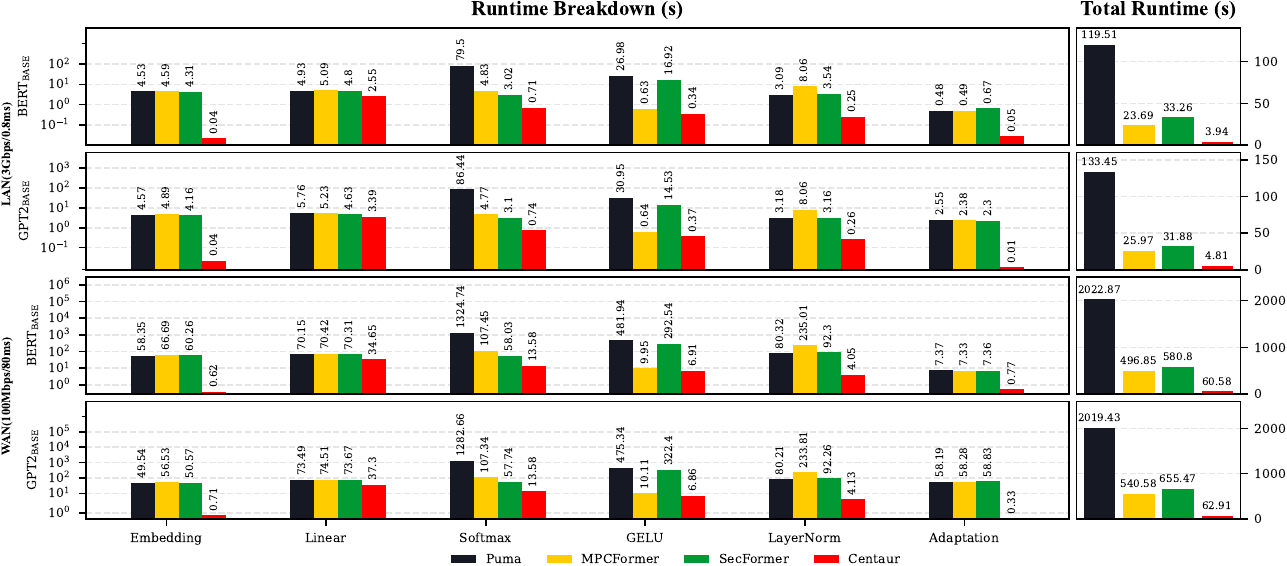} 
	\caption{Time breakdown for BERT$_{\text{BASE}}$ and GPT-2$_{\text{BASE}}$. The results are the average of ten runs.}
\label{fig: time_breakdown_base}
\end{figure*}

\noindent\textbf{Embedding \& Adaptation Layers.} 
The Embedding and Adaptation layers both include linear and nonlinear operations, allowing \textsc{Centaur} to achieve dual optimization in communication overhead. Specifically, for the Embedding layer, which includes matrix multiplication and LayerNorm, \textsc{Centaur} reduces communication overhead by 22.0 $\sim$27.8 times compared to the current state-of-the-art PPTI frameworks. For the Adaptation layer, \textsc{Centaur} reduces communication overhead by $10.2$ and $11.2$ times on the BERT series models. However, for the GPT-2 series models, the reductions are significantly higher, at $448.3$ and $698.7$ times. This is due to the different structures used in the adaptation layers of BERT and GPT-2 models to adapt to downstream tasks.

\subsection{More Time Breakdown Results}\label{sec: base_time_breakdown}
In this section, we present the results of the time overhead for privacy-preserving inference using \textsc{Centaur} with BERT$_{\text{BASE}}$ and GPT-2$_{\text{BASE}}$ models under LAN and WAN settings. The analysis results are consistent with those observed for BERT$_{\text{LARGE}}$ and GPT-2$_{\text{LARGE}}$ \cref{sec: performance}.

\begin{table*}[ht]
\centering
\caption{The cost of privacy-preserving inference on LLaMA-7B, where \#Input denotes the length of the input sentence and \#Output represents the number of generated tokens.}
\label{tab: llama}
\begin{tabular}{|c|c|c|c|c|c|c|}
\hline
(\#Input, \#Output) & \multicolumn{2}{c|}{(4, 1)} & \multicolumn{2}{c|}{(8, 1)} & \multicolumn{2}{c|}{(16, 1)} \\ \hline
Costs & Comm(GB) & Time(S) & Comm(GB) & Time(S) & Comm(GB) & Time(S) \\ \hline
\textsc{Centaur} & 0.32 & 2.76 & 0.39 & 3.02 & 0.54 & 6.81 \\ \hline
\end{tabular}
\end{table*}

\section{The Generalizability of \textsc{Centaur}} \label{sec: generalizability}
\textsc{Centaur} is compatible with other Transformer models and can achieve a more optimal balance between privacy, efficiency, and performance. This is due to \textsc{Centaur} performing the computation of nonlinear functions in Transformer models under permutation, allowing for seamless extension to other Transformer architectures, such as LLaMA. In particular, the LLaMA model utilizes the RMSNorm normalization function and the SwiGLU activation function. These functions are analogous to LayerNorm and GeLU and can both be computed under permutation, as demonstrated below:

\begin{equation}
    \text{RMSNorm}(\mathbf{x \pi}) = \frac{\mathbf{x\pi}}{\sqrt{\frac{1}{d} \sum \limits_{i=1}^{d}} x_i^2} = \text{RMSNorm}(\mathbf{x})\pi,
\end{equation}

\begin{equation}
    \text{SwiGLU}(\mathbf{x} \pi) = \frac{\mathbf{x} \pi}{1 + e^{-\mathbf{x}\pi}} = \text{SwiGLU}(x)\pi.
\end{equation}

Where $\mathbf{x} \in \mathbb{R}^d$ is the input vector, $d$ is the input dimension (i.e., the number of elements), and $x_i$ represents the $i$-th element in the vector. This enables \textsc{Centaur} to perform complete privacy-preserving inference on the LLaMA model without changing its underlying architecture. In contrast, for SMPC-based PPTI frameworks, such as MPCFormer and PUMA, extending to the LLaMA model would require the design of proprietary SMPC protocols for handling RMSNorm and SwiGLU. This means that \textsc{Centaur} is more generalizable than SMPC-based PPTI frameworks. 

Since \textsc{Centaur} does not alter the structure of the LLaMA model, it can theoretically achieve performance comparable to the plaintext model. In terms of efficiency, we have further added experimental results of \textsc{Centaur} applied to the LLaMA-7B model. We used the same experimental setup as in the paper and executed the experiments in a local area network (LAN) with 20Gbps bandwidth and 0.1ms latency. 

From the data in \cref{tab: llama}, it is evident that \textsc{Centaur} can complete privacy-preserving inference on the LLaMA-7B model in less than 10 seconds, with communication overheads below 1GB. When the input sequence length is 8, executing privacy-preserving inference on the LLaMA-7B model using \textsc{Centaur} generates 1 token in less than 3 seconds, with a communication overhead of 0.39GB. In the same network conditions (bandwidth and latency), PUMA would require approximately 200 seconds and 1.79GB of communication. This shows that \textsc{Centaur} offers significant advantages in both speed and communication efficiency, making it a highly scalable and practical solution for privacy-preserving Transformer model inference.

\section{Hyper-parameter}\label{sec: hyper-parameter}
For the baselines MPCFormer \cite{li2022mpcformer} and SecFormer \cite{luo2024secformer}, which require additional training and distillation, we followed the fine-tuning and distillation hyperparameter selection method as described in \cite{li2022mpcformer}. Specifically, for BERT series models, during the fine-tuning phase, we used learning rates of \{1e-6, 5e-6, 1e-5, 1e-4\}, batch sizes of \{64, 256\}, and epochs of \{10, 30, 100\}. For GPT-2 series models, during the fine-tuning phase, we used learning rates of \{1e-6, 5e-6, 1e-5, 1e-4\}, a batch size of 2, and epochs of \{1, 3, 5\}. We fine-tuned each model with these hyperparameter combinations and selected the best-performing model as the teacher. 

During the knowledge distillation phase, for BERT series models, the number of distillation iterations was determined based on the MSE loss between the embedding layer and the transformer layer. For small datasets (CoLA, MRPC, RTE), the batch size was 8, while for large datasets (QNLI, STS-B), the batch size was 32. Specifically, for the distillation stages in the embedding layer and transformer layer, QNLI was trained for 10 epochs, MRPC for 20 epochs, STS-B for 50 epochs, CoLA for 50 epochs, and RTE for 50 epochs. For GPT-2 models, we used KLDiv loss to calculate the loss between the output representations of the teacher and student models, and Cosine loss to calculate the loss between the hidden layers of the teacher and student models. The number of distillation steps was determined based on the loss values.

\end{document}